\documentclass[letterpaper]{article} 
\usepackage{aaai2027}

\usepackage[hyphens]{url} 
\usepackage{graphicx} 
\usepackage{natbib} 
\usepackage{caption} 
\usepackage{amsmath}
\usepackage{amsfonts}
\usepackage{amssymb}
\usepackage{xcolor}
\usepackage{booktabs}
\usepackage{multirow}

\definecolor{bestbg}{gray}{0.88}

\newcommand{\bestresult}[1]{%
  \begingroup
  \setlength{\fboxsep}{0.7pt}%
  \colorbox{bestbg}{\textbf{#1}}%
  \endgroup
}

\newcommand{\secondresult}[1]{\textbf{#1}}

\title{STEK: Structural-Terrain-Enhanced Koopman-Inspired Cross-Channel Regularization for Retrospective Battery Multi-Target Estimation}

\author{
    Jing Wang\textsuperscript{\rm 1},
    Shiyuan Piao\textsuperscript{\rm 2},
    Fugee Tsung\textsuperscript{\rm 2}
}
\affiliations{
    \textsuperscript{\rm 1}University of Science and Technology of China, Hefei, China\\
    \textsuperscript{\rm 2}The Hong Kong University of Science and Technology, Hong Kong, China\\
    jingwang11@mail.ustc.edu.cn, shiyuan.piao@connect.ust.hk, season@ust.hk
}

\begin{document}

\maketitle

\begin{abstract}
Retrospective battery characterization across heterogeneous cells is
challenging because cycling records vary substantially with chemistry,
protocol, cell format, capacity, and operating conditions. An endpoint
charge--discharge curve captures the local condition of an evaluated
cycle, while a cell-level lifecycle terrain summarizes structural
evolution across the archived trajectory.
We propose STEK, a structural-terrain-enhanced framework for
retrospective battery multi-target estimation. STEK combines a
cell-level lifecycle \(dQ/dV\) terrain with an endpoint-derived
structural proxy through quality-proxy-conditioned gating. Bounded
FiLM and a validity-gated residual head refine the endpoint-curve
representation while preserving a curve-based prediction pathway.
Battery-type and optional material-text representations further
condition an identity-centered low-rank linear map, providing a
Koopman-inspired cross-channel consistency objective without modeling
temporal dynamics or performing rollout.
Under the fixed seven-output MIX-battery protocol, the full
retrospective setting improves aggregate performance over
endpoint-only learned baselines across multiple temporal
backbones; this comparison reflects both the additional
archived context and the modeling design. Under matched
auxiliary inputs, STEK outperforms direct concatenation and
quality-unaware FiLM, while descriptor-removal tests show
graceful fallback to the endpoint pathway. Deterministic
rules remain preferable for directly recoverable capacity
summaries. The method targets retrospective archive
harmonization rather than online estimation from cycle $t$
alone.
\end{abstract}

\section{Introduction}

Retrospective cycle-level characterization of heterogeneous batteries
from archived cycling records is challenging because cycling curves
vary across chemistries, cell formats, nominal capacities,
temperatures, and operating protocols. These factors entangle
degradation-related information with source-specific variation,
limiting the transferability of curve-only representations across
battery sources~\cite{36,lu2023deep,tao2024generative}.

Deep sequence models learn nonlinear representations directly from
voltage, current, and capacity curves. Transformer-based and
frequency-aware architectures have improved the modeling of complex
and non-stationary signal variation
~\cite{46,zhou2022fedformer,56}. However, an endpoint curve alone does
not summarize structural evolution across an archived cycling
trajectory or account for differences in material context.
Incremental-capacity structure provides a complementary view of
degradation, while material descriptors help distinguish cells with
different recorded chemistries. Existing approaches often omit such
information or incorporate it through direct concatenation, which
provides limited control when auxiliary descriptors are missing or
unreliable.

Koopman-inspired representation learning introduces linear structure
in a lifted observable space. Existing methods primarily learn temporal
state transitions or support dynamical rollout
~\cite{knf2023,skolr2025,koonpro2025,vitkoop2025}. STEK instead uses an
identity-centered, sample-conditioned low-rank linear map to regularize
lifted voltage, current, and capacity representations within the same
endpoint curve. Their ordering specifies channel roles rather than
successive physical states. The resulting cross-channel constraint is
used only during training and is not intended to identify battery
dynamics, forecast future cycles, or perform dynamical rollout.

We propose STEK, a structural-terrain-enhanced framework for
retrospective battery multi-target estimation. For each evaluated
endpoint, STEK combines an endpoint-local charge--discharge curve with
a cell-level (dQ/dV) lifecycle terrain constructed from the archived
trajectory and an endpoint-derived structural proxy.
Quality-proxy-conditioned fusion and bounded FiLM conservatively
modulate the curve representation, while a validity-gated residual head
retains the curve-based prediction as a fallback. Battery-type and
optional material-text representations
~\cite{molt5,foundationbattery2025} further condition the
Koopman-inspired cross-channel regularizer. Because the lifecycle terrain may include cycles after endpoint
$t$, STEK targets retrospective archive harmonization rather
than online or prospective prediction. It is intended for
heterogeneous collections in which standardized curves are
retained but derived annotations or some target-defining raw
fields are missing or inconsistent. When all raw signals required
by a target definition are available, direct analytic computation
remains preferable.
Our contributions are as follows:

\begin{itemize}

\item We formulate retrospective battery multi-target estimation by
combining endpoint-local charge--discharge curves with archived
cell-level lifecycle structure. We develop a conservative enhancement
mechanism based on quality-proxy-conditioned fusion, bounded FiLM, and
validity-gated residual prediction while preserving an endpoint-curve
fallback.

\item We introduce a context-conditioned lifted cross-channel
regularizer based on an identity-centered low-rank linear map
with residual corrections. The design is motivated by
Koopman-style linear consistency in a lifted space, but it is
used only as a training-time representation constraint and
does not identify physical dynamics or require rollout.

\item We evaluate STEK on the ten-source MIX-battery benchmark using
endpoint-only baselines, matched-input fusion controls, multiple
temporal backbones, component ablations, cross-source transfer on three
held-out sources, descriptor-availability stress tests, and efficiency
analysis. Input-recoverability diagnostics further distinguish
deterministic capacity summaries from history-dependent and
cross-signal estimation targets.

\end{itemize}

\section{Related Work}

\subsection{Deep Battery Modeling}

Battery modeling has advanced from generic sequence fitting to
specialized temporal, patch-based, frequency-aware, and state-space
architectures~\cite{43,44,46,45,zeng2023dlinear,zhou2022fedformer,
koopa2023,56,shi2024mambalithium}. Foundation models further improve
transfer across tasks, chemistries, and operating conditions
~\cite{53,54,56,57}. However, most methods rely primarily on endpoint
curves and do not explicitly integrate archived lifecycle structure or control the use of auxiliary material information.

\subsection{Koopman-Based Representation Learning}

Koopman theory describes nonlinear dynamics through linear evolution
in a lifted space~\cite{koopman1931}. Neural extensions support
distribution shifts, structured recurrence, probabilistic dynamics,
and spatiotemporal forecasting
~\cite{knf2023,skolr2025,koonpro2025,vitkoop2025}. In contrast, STEK
uses an identity-centered low-rank linear component as a
Koopman-inspired consistency constraint across lifted voltage,
current, and capacity representations, without interpreting them as
temporal states or performing rollout.

\subsection{Electrochemical and Multimodal Battery Modeling}

Incremental-capacity, differential-voltage, and impedance-related
descriptors provide complementary degradation information, while
physics-informed methods incorporate electrochemical knowledge
~\cite{pinn_soh2026}. Molecular and language models further support
chemical and material-text representations
~\cite{liu2022graphmvp,xia2023molebert,zhou2023unimol,molt5,
foundationbattery2025}. Unlike direct concatenation
~\cite{liang2021multibench}, STEK combines quality-proxy-conditioned
fusion, bounded feature modulation, and validity-gated correction
while retaining an endpoint-curve pathway.

\begin{figure}[htbp]
    \centering
    \includegraphics[width=0.95\columnwidth]{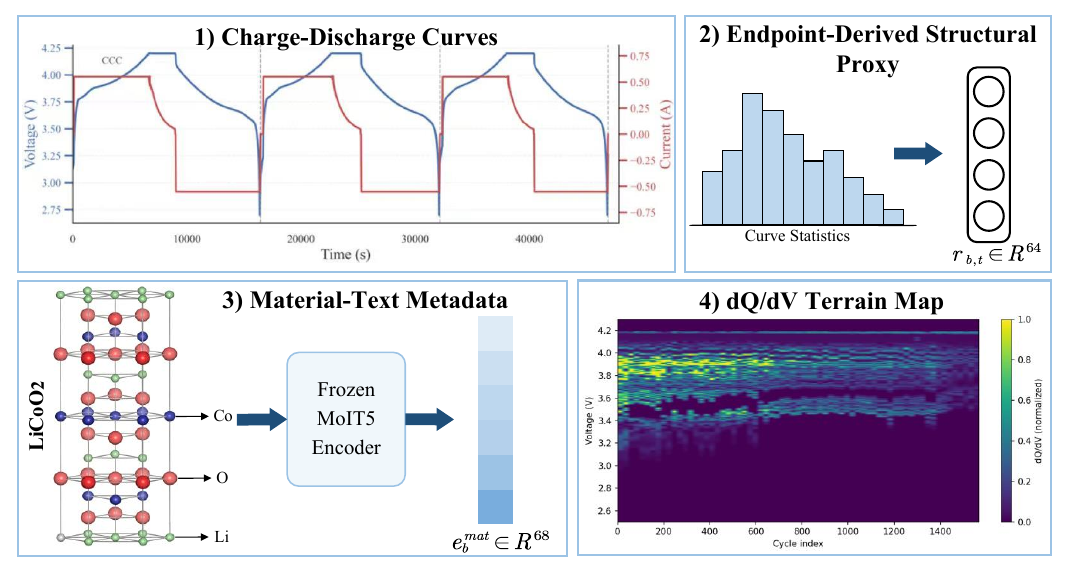}
    \caption{Endpoint and lifecycle inputs used by STEK.}
    \label{fig:framework_input}
\end{figure}

\begin{figure*}[htbp]
    \centering
    \includegraphics[width=0.95\textwidth]{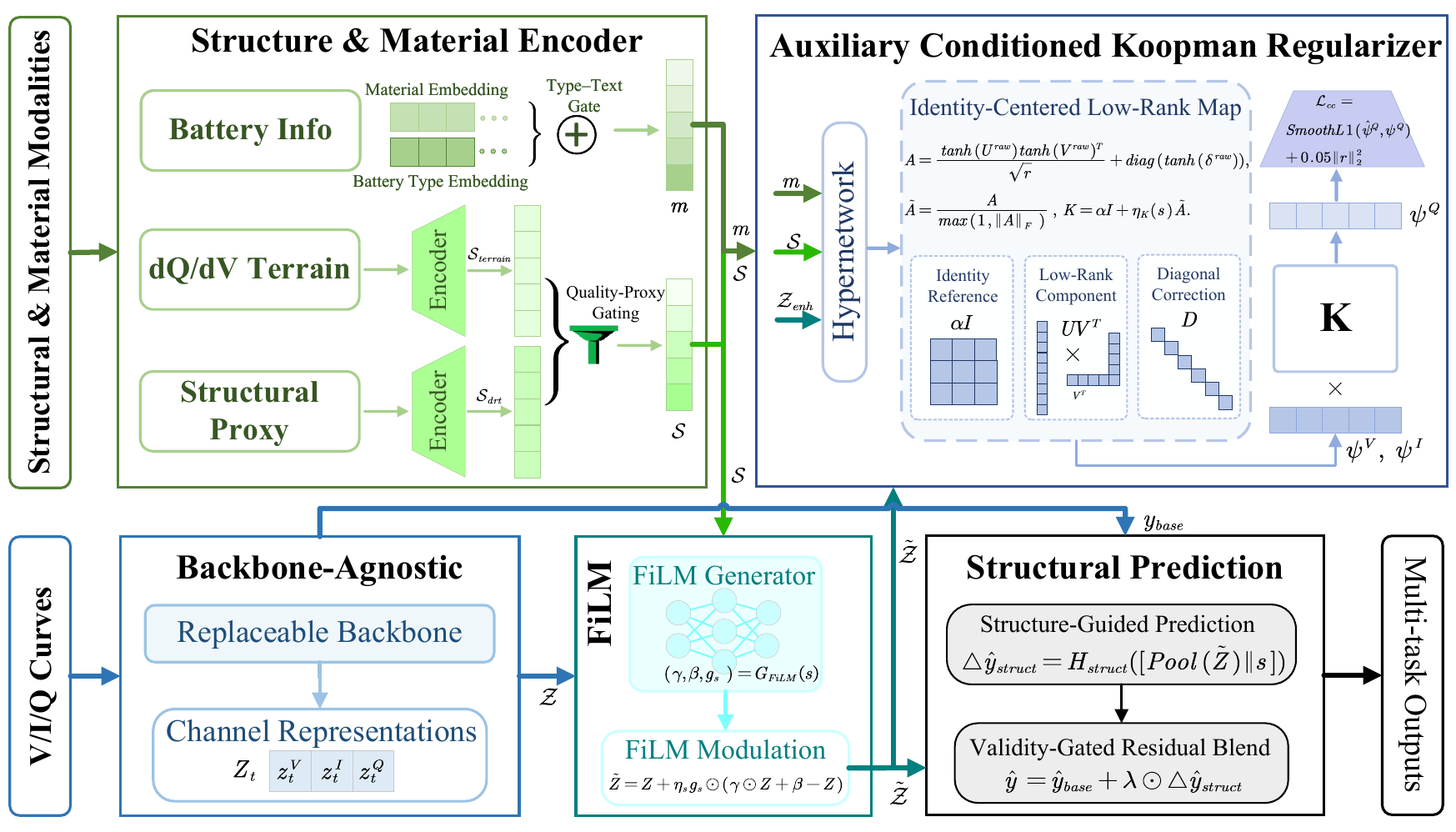}
    \caption{STEK architecture. Lifecycle context refines endpoint
    predictions, while structural and material context condition a
    training-only Koopman-inspired cross-channel regularizer.}
    \label{fig:arch}
\end{figure*}

\section{Method}
\label{sec:method}

\subsection{Problem Formulation and Overview}

Let
\[
\mathcal{X}_b
=
\{X_{b,\tau}\}_{\tau=1}^{N_b}
\]
denote the archived trajectory of cell \(b\). At endpoint \(t\), the
local input contains resampled voltage, current, and capacity traces:
\begin{equation}
X_{b,t}
=
[X_{b,t}^{V};X_{b,t}^{I};X_{b,t}^{Q}]
\in\mathbb{R}^{3\times L},
\qquad L=100.
\label{eq:curve_input}
\end{equation}
The target \(\mathbf{y}_{b,t}\) contains cycle-level attributes, some
defined using historical references.

STEK also uses a cell-level lifecycle terrain
\begin{equation}
\mathbf{T}_b^{\mathrm{life}}
=
f_{\mathrm{terrain}}(\mathcal{X}_b),
\label{eq:lifecycle_terrain}
\end{equation}
an endpoint-derived structural proxy \(\mathbf{r}_{b,t}\), battery
type \(c_b\), and an optional material-text representation
\(e_b^{\mathrm{mat}}\). Their availability is denoted by
\[
\mathbf{v}_{b,t}
=
[v_{T,b},v_{R,b,t},v_{m,b}]
\in\{0,1\}^{3}.
\]
The terrain is shared across endpoints of cell \(b\). This defines a
retrospective lifecycle-context setting, not prospective estimation
restricted to observations available by endpoint \(t\).

As shown in Figs.~\ref{fig:framework_input} and~\ref{fig:arch}, a
replaceable backbone extracts one representation per endpoint channel:
\begin{equation}
Z_{b,t}
=
\mathcal{B}_{\theta}(X_{b,t})
=
[z_{b,t}^{V},z_{b,t}^{I},z_{b,t}^{Q}]
\in\mathbb{R}^{3\times d}.
\label{eq:backbone}
\end{equation}
Structural context modulates \(Z_{b,t}\) and supports residual
prediction. Structural and material context also condition a
training-only cross-channel regularizer.

\subsection{Quality-Proxy Structural Encoding}

The lifecycle terrain and endpoint-derived structural proxy are encoded independently:
\begin{equation}
F_{T,b}
=
E_T(\mathbf{T}_b^{\mathrm{life}}),
\qquad
F_{R,b,t}
=
E_R(\mathbf{r}_{b,t}).
\label{eq:structural_encoding}
\end{equation}
The structural proxy summarizes endpoint voltage, current, capacity,
their first differences, and incremental-capacity shape; details are
provided in the supplementary material.

Descriptor statistics \(\xi\) summarize magnitude, variation, and
valid-observation ratio. The quality-proxy gates are
\begin{equation}
\begin{aligned}
q_{T,b}
&=
v_{T,b}\,
\sigma\!\left(
Q_T([F_{T,b}\Vert\xi_{T,b}])
\right),\\
q_{R,b,t}
&=
v_{R,b,t}\,
\sigma\!\left(
Q_R([F_{R,b,t}\Vert\xi_{R,b,t}])
\right).
\end{aligned}
\label{eq:quality_gates}
\end{equation}
The fused structural state is
\begin{equation}
s_{b,t}
=
F_s\!\left(
[q_{T,b}F_{T,b}
\Vert q_{R,b,t}F_{R,b,t}
\Vert q_{T,b}
\Vert q_{R,b,t}
\Vert v_{T,b}
\Vert v_{R,b,t}]
\right).
\label{eq:struct_fusion}
\end{equation}
Availability masks suppress missing inputs, while the learned gates
weight available descriptors. The gates are not interpreted as
calibrated confidence estimates.

\subsection{Material Context Encoding}

Material context combines battery type with an optional
metadata-derived material-text representation:
\begin{equation}
\begin{aligned}
e_b^{\mathrm{type}}
&=
\operatorname{Embed}(c_b),\\
e_b^{\mathrm{text}}
&=
P_m(e_b^{\mathrm{mat}}),\\
m_b
&=
e_b^{\mathrm{type}}
+
v_{m,b}\sigma(w_m)\odot
\left(
e_b^{\mathrm{text}}-e_b^{\mathrm{type}}
\right).
\end{aligned}
\label{eq:material_gate}
\end{equation}
When material text is unavailable,
\(m_b=e_b^{\mathrm{type}}\). These representations encode available recorded metadata and
may inherit source-level labels; they are not interpreted as
verified cell-specific molecular structures. Material context
conditions only the training-time regularizer and does not
directly enter the prediction head.

\subsection{Bounded Structural Modulation}

The structural state modulates each channel through bounded residual
FiLM:
\begin{equation}
\begin{aligned}
\gamma_{b,t}
&=
1+\delta_\gamma\tanh(G_\gamma(s_{b,t})),\\
\beta_{b,t}
&=
\delta_\beta\tanh(G_\beta(s_{b,t})),\\
a_{b,t}
&=
\eta_s\sigma\!\left(
G_s([s_{b,t}\Vert q_{T,b}\Vert q_{R,b,t}
\Vert v_{T,b}\Vert v_{R,b,t}])
\right),\\
\tilde{z}_{b,t}^{c}
&=
z_{b,t}^{c}
+
a_{b,t}\odot
\left[
(\gamma_{b,t}-1)\odot z_{b,t}^{c}
+\beta_{b,t}
\right],
\quad c\in\{V,I,Q\}.
\end{aligned}
\label{eq:bounded_film}
\end{equation}
Here, \(\delta_\gamma\) and \(\delta_\beta\) bound the feature-wise
scale and shift, while \(0\leq\eta_s\leq1\) limits the correction.
The endpoint-curve representation remains the reference.

\subsection{Lifted Cross-Channel Consistency Regularization}

Motivated by Koopman-style linear consistency, each modulated
channel is mapped into a lifted observable space:
\begin{equation}
\psi_{b,t}^{c}
=
\mathcal{E}_{\mathrm{lift}}(\tilde{z}_{b,t}^{c}),
\qquad c\in\{V,I,Q\}.
\label{eq:lifting}
\end{equation}
The regularizer context is formed by concatenating the material,
structural, and pooled curve representations, followed by layer
normalization:
\begin{equation}
c_{K,b,t}
=
\operatorname{LN}\!\left(
[m_b \Vert s_{b,t} \Vert \operatorname{Pool}(\widetilde Z_{b,t})]
\right).
\label{eq:koopman_context}
\end{equation}

Three projection heads generate the raw low-rank and diagonal
components:
\begin{equation}
\begin{aligned}
U_{b,t}^{\mathrm{raw}} &= \Phi_U(c_{K,b,t}),\\
V_{b,t}^{\mathrm{raw}} &= \Phi_V(c_{K,b,t}),\\
\delta_{b,t}^{\mathrm{raw}} &= \Phi_D(c_{K,b,t}).
\end{aligned}
\end{equation}
The bounded perturbation and sample-conditioned update scale are
\begin{equation}
\begin{aligned}
A_{b,t}
&=
\frac{
\tanh(U_{b,t}^{\mathrm{raw}})
\tanh(V_{b,t}^{\mathrm{raw}})^\top
}{\sqrt r}
+
\operatorname{diag}\!\left(
\tanh(\delta_{b,t}^{\mathrm{raw}})
\right),\\
\widetilde A_{b,t}
&=
\frac{A_{b,t}}
{\max\!\left(1,\lVert A_{b,t}\rVert_F\right)},\\
g_{b,t}
&=
\sigma\!\left(G_{\eta}(s_{b,t})\right),\\
\eta_{K,b,t}
&=
\eta_0\left(0.25+0.75g_{b,t}\right),\\
K_{b,t}
&=
\alpha I+\eta_{K,b,t}\widetilde A_{b,t}.
\end{aligned}
\label{eq:conditioned_operator}
\end{equation}
Here, $U_{b,t}^{\mathrm{raw}},V_{b,t}^{\mathrm{raw}}
\in\mathbb{R}^{d_K\times r}$ and
$\delta_{b,t}^{\mathrm{raw}}\in\mathbb{R}^{d_K}$.
We use $\eta_0=0.05$ and $\alpha=0.90$, so
$\eta_{K,b,t}\in(0.0125,0.05)$. Since
$\|\widetilde A_{b,t}\|_2\leq1$, the construction gives
$\|K_{b,t}\|_2<0.95$, limiting amplification without implying
physical dynamics or rollout stability.

Using the implementation ordering \((V,I,Q)\), define
\begin{equation}
\psi_{b,t}^{-}:=\psi_{b,t}^{V},\qquad
\psi_{b,t}^{0}:=\psi_{b,t}^{I},\qquad
\psi_{b,t}^{+}:=\psi_{b,t}^{Q}.
\label{eq:channel_roles}
\end{equation}
These symbols denote channel roles rather than temporal states. The
capacity-channel observable is reconstructed as
\begin{equation}
\begin{aligned}
\epsilon_{b,t}^{K}
&=
\eta_r\tanh\!\left(
R(\operatorname{sg}(\psi_{b,t}^{0}))
\right),\\
\widehat{\psi}_{b,t}^{+}
&=
d_{b,t}K_{b,t}\psi_{b,t}^{0}
+
m_{K,b,t}\odot\psi_{b,t}^{-}
+
\epsilon_{b,t}^{K}.
\end{aligned}
\label{eq:observable_prediction}
\end{equation}
Here, \(\operatorname{sg}(\cdot)\) denotes stop-gradient.
Separate two-layer MLP heads generate
\(d_{b,t}\in(0.90,1.10)\) and
\(m_{K,b,t}\in(0,0.05)^{d_K}\) from \(c_{K,b,t}\), and
\(\eta_r=0.05\). Exact parameterizations are given in the
supplementary material. Cross-channel consistency is imposed by
\begin{equation}
\mathcal{L}_{\mathrm{cc}}
=
\operatorname{SmoothL1}
\left(
\widehat{\psi}_{b,t}^{+},
\psi_{b,t}^{+}
\right)
+
0.05\left\|
\epsilon_{b,t}^{K}
\right\|_2^2.
\label{eq:koopman_consistency}
\end{equation}
This is a within-curve representation constraint rather than a
cycle-transition or dynamical-rollout objective. Accordingly,
$K_{b,t}$ is not interpreted as a shared physical Koopman
operator.

\subsection{Structural Prediction and Residual Blending}

The endpoint prediction and structural residual are
\begin{equation}
\begin{aligned}
\hat{\mathbf{y}}_{b,t}^{\mathrm{base}}
&=
H_{\mathrm{base}}
\left(
\operatorname{Pool}(Z_{b,t})
\right),\\
\Delta\hat{\mathbf{y}}_{b,t}^{\mathrm{struct}}
&=
H_{\mathrm{struct}}\!\left(
[\operatorname{Pool}(\tilde{Z}_{b,t})\Vert s_{b,t}]
\right).
\end{aligned}
\label{eq:prediction_heads}
\end{equation}
A bounded gate controls the structural residual:
\begin{equation}
\begin{aligned}
\lambda_{b,t}
&=
\lambda_{\max}\sigma\!\left(
G_\lambda(
[s_{b,t}\Vert q_{T,b}\Vert q_{R,b,t}
\Vert v_{T,b}\Vert v_{R,b,t}]
)
\right),\\
\hat{\mathbf{y}}_{b,t}
&=
\hat{\mathbf{y}}_{b,t}^{\mathrm{base}}
+
\lambda_{b,t}\odot
\Delta\hat{\mathbf{y}}_{b,t}^{\mathrm{struct}}.
\end{aligned}
\label{eq:prediction_blending}
\end{equation}
The gate uses descriptor-quality proxies and availability indicators,
while retaining the endpoint estimate as the reference.

The complete training objective is
\begin{equation}
\mathcal{L}
=
\mathcal{L}_{\mathrm{task}}
+\lambda_{\mathrm{cc}}\mathcal{L}_{\mathrm{cc}}.
\label{eq:overall_loss}
\end{equation}
We set \(\lambda_{\mathrm{cc}}=0.05\).
Here, \(\mathcal{L}_{\mathrm{task}}\) is detailed in the supplementary
material, and \(\mathcal{L}_{\mathrm{cc}}\) is used only during training.

\section{Experiment}

\subsection{Datasets}
MIX-battery~\cite{33,34,35,31,19,37,38,39,3,41} contains 511
candidate cells from ten battery-degradation sources. Cell-disjoint
training, validation, and test splits contain 326/92/93 candidate
cells. Endpoint filtering within each split retains 302/86/84 cells
and 1,501/426/414 samples, respectively, for a total of 472 cells and
2,341 samples. All endpoints from a cell remain in the same split.
The other 39 cells lack a usable end-of-life record or an eligible
endpoint at or beyond cycle 96. Further dataset and eligibility
details are provided in the supplementary material.

\begin{table*}[t]
\centering
\small
\setlength{\tabcolsep}{2.5pt}
\renewcommand{\arraystretch}{0.95}
\begin{tabular}{llccccccc}
\toprule
\textbf{Task} & \textbf{Metric}
& \textbf{STEK}
& \textbf{MambaLithium}
& \textbf{LiPM}
& \textbf{CPTransformer}
& \textbf{iTransformer}
& \textbf{Koopa}
& \textbf{KNF} \\
\midrule

\multirow{2}{*}{NDC}
& MAE $\downarrow$
& \bestresult{0.90 $\pm$ 0.06}
& 1.55 $\pm$ 0.32
& 2.07 $\pm$ 0.22
& 1.26 $\pm$ 0.03
& \secondresult{1.16 $\pm$ 0.07}
& 1.25 $\pm$ 0.12
& 1.35 $\pm$ 0.06 \\

& MaxAE $\downarrow$
& \bestresult{7.65 $\pm$ 1.15}
& 13.27 $\pm$ 0.74
& 16.16 $\pm$ 8.26
& 13.62 $\pm$ 1.42
& 12.04 $\pm$ 1.62
& 10.85 $\pm$ 3.98
& \secondresult{9.34 $\pm$ 0.49} \\

\multirow{2}{*}{SOH}
& MAE $\downarrow$
& \bestresult{0.40 $\pm$ 0.03}
& 0.56 $\pm$ 0.02
& 0.65 $\pm$ 0.03
& 0.60 $\pm$ 0.06
& \secondresult{0.55 $\pm$ 0.02}
& 0.62 $\pm$ 0.05
& 0.63 $\pm$ 0.02 \\

& RMSE $\downarrow$
& \bestresult{0.99 $\pm$ 0.02}
& 1.51 $\pm$ 0.02
& \secondresult{1.21 $\pm$ 0.03}
& 1.50 $\pm$ 0.12
& 1.46 $\pm$ 0.07
& 1.44 $\pm$ 0.14
& 1.33 $\pm$ 0.04 \\

\multirow{2}{*}{CF}
& MAE $\downarrow$
& \bestresult{0.38 $\pm$ 0.02}
& 0.57 $\pm$ 0.02
& 0.65 $\pm$ 0.06
& 0.60 $\pm$ 0.06
& \secondresult{0.55 $\pm$ 0.03}
& 0.62 $\pm$ 0.04
& 0.64 $\pm$ 0.03 \\

& RMSE $\downarrow$
& \bestresult{0.97 $\pm$ 0.02}
& 1.51 $\pm$ 0.01
& \secondresult{1.20 $\pm$ 0.03}
& 1.50 $\pm$ 0.12
& 1.47 $\pm$ 0.08
& 1.44 $\pm$ 0.13
& 1.34 $\pm$ 0.03 \\

\multirow{2}{*}{CFR}
& MAE $\downarrow$
& \secondresult{1.33 $\pm$ 0.00}
& 1.35 $\pm$ 0.03
& 1.50 $\pm$ 0.01
& 1.49 $\pm$ 0.11
& \secondresult{1.33 $\pm$ 0.01}
& \bestresult{1.31 $\pm$ 0.01}
& 1.40 $\pm$ 0.01 \\

& RMSE $\downarrow$
& \secondresult{3.00 $\pm$ 0.05}
& 3.05 $\pm$ 0.05
& 3.17 $\pm$ 0.02
& 3.36 $\pm$ 0.26
& 3.02 $\pm$ 0.05
& \bestresult{2.93 $\pm$ 0.05}
& \secondresult{3.00 $\pm$ 0.02} \\

\multirow{2}{*}{ICR}
& MAE ($\mathrm{m}\Omega$) $\downarrow$
& \bestresult{6.54 $\pm$ 0.98}
& 10.42 $\pm$ 0.35
& 16.81 $\pm$ 2.17
& \secondresult{8.11 $\pm$ 0.25}
& 9.00 $\pm$ 0.64
& 10.79 $\pm$ 0.95
& 11.06 $\pm$ 0.49 \\

& RE (\%) $\downarrow$
& \bestresult{11.99 $\pm$ 2.45}
& 20.46 $\pm$ 0.49
& 37.47 $\pm$ 5.10
& 19.48 $\pm$ 1.79
& \secondresult{18.88 $\pm$ 0.69}
& 28.82 $\pm$ 7.74
& 25.86 $\pm$ 4.71 \\

\multirow{2}{*}{NCC}
& MAE $\downarrow$
& \bestresult{0.85 $\pm$ 0.03}
& 1.37 $\pm$ 0.21
& 2.01 $\pm$ 0.18
& 1.20 $\pm$ 0.09
& \secondresult{1.06 $\pm$ 0.06}
& 1.17 $\pm$ 0.10
& 1.35 $\pm$ 0.15 \\

& RMSE $\downarrow$
& \bestresult{1.31 $\pm$ 0.09}
& 2.03 $\pm$ 0.18
& 3.24 $\pm$ 0.52
& 1.96 $\pm$ 0.10
& 1.77 $\pm$ 0.01
& \secondresult{1.76 $\pm$ 0.23}
& 2.08 $\pm$ 0.28 \\

QTY
& RQE (\%) $\downarrow$
& \bestresult{6.74 $\pm$ 0.64}
& \secondresult{17.24 $\pm$ 0.50}
& 66.67 $\pm$ 10.29
& 19.43 $\pm$ 8.32
& 17.91 $\pm$ 1.67
& 23.27 $\pm$ 6.13
& 28.18 $\pm$ 5.62 \\

\midrule
OVERALL
& Score $\downarrow$
& \bestresult{12.42 $\pm$ 0.30}
& 16.24 $\pm$ 0.59
& 22.11 $\pm$ 1.10
& 16.51 $\pm$ 1.29
& \secondresult{15.08 $\pm$ 0.47}
& 16.27 $\pm$ 0.45
& 17.26 $\pm$ 0.38 \\
\bottomrule
\end{tabular}
\caption{Performance under the fixed seven-output MIX-battery
protocol. STEK uses endpoint curves and archived auxiliary context;
the other learned models are endpoint-only. Thus, the table compares
complete settings rather than matched-input architectures;
Table~\ref{tab:targeted_design_analysis} provides matched-input
controls. NDC, SOH, CF, NCC, and Overall are multiplied by $10^2$,
and CFR by $10^4$. Deterministic NDC/NCC rules are reported in the
supplementary material. Lower values are better. }
\label{tab:mix_battery_comparison}
\end{table*}

\begin{figure}[t]
\centering
\includegraphics[width=0.95\columnwidth]{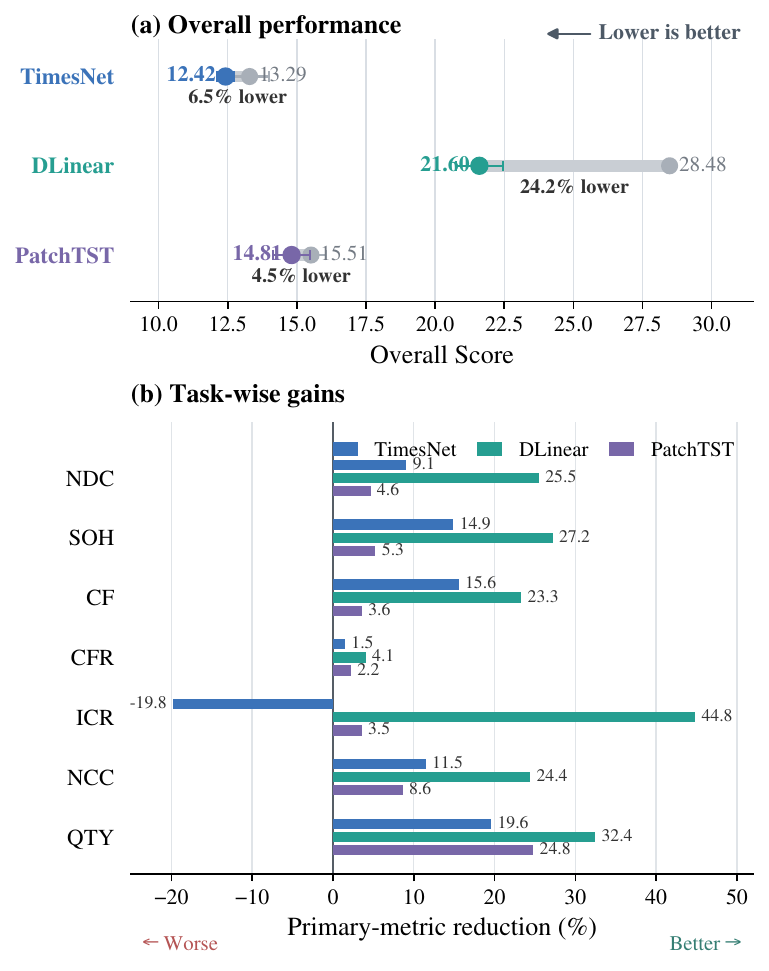}
\caption{
Generality of STEK across temporal backbones.
(a) Overall scores before and after adding STEK.
(b) Relative improvement on each output's primary metric.
}
\label{fig:backbone_generality}
\end{figure}

\begin{table*}[!t]
\centering
\small
\setlength{\tabcolsep}{1.9pt}
\renewcommand{\arraystretch}{0.95}

\begin{tabular}{llc|ccc|ccc}
\toprule
& & \multirow{2}{*}{\textbf{Full STEK}}
& \multicolumn{3}{c|}{\textbf{Reduced Modality Sets}}
& \multicolumn{3}{c}{\textbf{Module Ablations}} \\
\cmidrule(lr){4-6}
\cmidrule(lr){7-9}

\textbf{Task} & \textbf{Metric}
&
& \shortstack{\textbf{Curve+Material}\\\textbf{+$dQ/dV$}}
& \shortstack{\textbf{Curve}\\\textbf{+Material}}
& \shortstack{\textbf{Curve}\\\textbf{+struct. Proxy}}
& \shortstack{\textbf{w/o}\\\textbf{FiLM}}
& \shortstack{\textbf{w/o Struct.}\\\textbf{Head}}
& \shortstack{\textbf{w/o Koopman}\\\textbf{Reg.}}\\
\midrule

\multirow{2}{*}{NDC}
& MAE $\downarrow$
& \bestresult{0.90 $\pm$ 0.06}
& 0.94 $\pm$ 0.10
& 1.00 $\pm$ 0.17
& 1.01 $\pm$ 0.02
& 1.06 $\pm$ 0.03
& 0.94 $\pm$ 0.10
& \secondresult{0.94 $\pm$ 0.07} \\

& MaxAE $\downarrow$
& \bestresult{7.65 $\pm$ 1.15}
& 8.83 $\pm$ 1.14
& 9.27 $\pm$ 2.03
& \secondresult{7.71 $\pm$ 2.78}
& 8.01 $\pm$ 0.25
& 8.02 $\pm$ 1.29
& 7.97 $\pm$ 2.11 \\

\midrule

\multirow{2}{*}{SOH}
& MAE $\downarrow$
& \secondresult{0.40 $\pm$ 0.03}
& \secondresult{0.40 $\pm$ 0.03}
& \secondresult{0.40 $\pm$ 0.01}
& \bestresult{0.39 $\pm$ 0.01}
& 0.44 $\pm$ 0.04
& 0.44 $\pm$ 0.04
& 0.40 $\pm$ 0.04 \\

& RMSE $\downarrow$
& \secondresult{0.99 $\pm$ 0.02}
& \bestresult{0.96 $\pm$ 0.02}
& 1.06 $\pm$ 0.12
& \bestresult{0.96 $\pm$ 0.06}
& 1.06 $\pm$ 0.06
& 1.04 $\pm$ 0.09
& 1.03 $\pm$ 0.13 \\

\midrule

\multirow{2}{*}{CF}
& MAE $\downarrow$
& \bestresult{0.38 $\pm$ 0.02}
& \secondresult{0.39 $\pm$ 0.02}
& 0.40 $\pm$ 0.01
& \secondresult{0.39 $\pm$ 0.02}
& 0.42 $\pm$ 0.03
& 0.42 $\pm$ 0.04
& 0.40 $\pm$ 0.04 \\

& RMSE $\downarrow$
& 0.97 $\pm$ 0.02
& \bestresult{0.94 $\pm$ 0.04}
& 1.04 $\pm$ 0.11
& \secondresult{0.95 $\pm$ 0.05}
& 1.02 $\pm$ 0.05
& 1.01 $\pm$ 0.09
& 1.00 $\pm$ 0.10 \\

\midrule

\multirow{2}{*}{CFR}
& MAE $\downarrow$
& \bestresult{1.33 $\pm$ 0.00}
& \secondresult{1.35 $\pm$ 0.01}
& 1.35 $\pm$ 0.03
& 1.37 $\pm$ 0.02
& 1.35 $\pm$ 0.02
& \bestresult{1.33 $\pm$ 0.05}
& 1.37 $\pm$ 0.05 \\

& RMSE $\downarrow$
& \secondresult{3.00 $\pm$ 0.05}
& 3.04 $\pm$ 0.04
& 3.02 $\pm$ 0.11
& 3.09 $\pm$ 0.02
& 3.06 $\pm$ 0.07
& \bestresult{2.99 $\pm$ 0.15}
& 3.07 $\pm$ 0.09 \\

\midrule

\multirow{2}{*}{ICR}
& MAE ($\mathrm{m}\Omega$) $\downarrow$
& 6.54 $\pm$ 0.98
& \secondresult{6.27 $\pm$ 0.27}
& 6.89 $\pm$ 0.53
& 6.67 $\pm$ 0.71
& 7.21 $\pm$ 0.56
& 6.77 $\pm$ 0.88
& \bestresult{5.96 $\pm$ 0.31} \\

& RE (\%) $\downarrow$
& 11.99 $\pm$ 2.45
& \bestresult{10.73 $\pm$ 0.60}
& 12.76 $\pm$ 2.30
& 11.78 $\pm$ 1.26
& 12.75 $\pm$ 2.06
& 11.60 $\pm$ 0.49
& \secondresult{11.57 $\pm$ 1.92} \\

\midrule

\multirow{2}{*}{NCC}
& MAE $\downarrow$
& \bestresult{0.85 $\pm$ 0.03}
& 0.90 $\pm$ 0.07
& 0.92 $\pm$ 0.06
& \secondresult{0.87 $\pm$ 0.03}
& 0.94 $\pm$ 0.08
& 1.02 $\pm$ 0.18
& 0.92 $\pm$ 0.06 \\

& RMSE $\downarrow$
& \bestresult{1.31 $\pm$ 0.09}
& \secondresult{1.38 $\pm$ 0.09}
& 1.42 $\pm$ 0.03
& \bestresult{1.31 $\pm$ 0.06}
& 1.44 $\pm$ 0.09
& 1.43 $\pm$ 0.18
& 1.40 $\pm$ 0.18 \\

\midrule

QTY
& RQE (\%) $\downarrow$
& \secondresult{6.74 $\pm$ 0.64}
& 10.81 $\pm$ 1.47
& 10.17 $\pm$ 1.44
& 7.69 $\pm$ 0.89
& \bestresult{5.92 $\pm$ 1.56}
& 9.34 $\pm$ 3.48
& 9.58 $\pm$ 2.64 \\

\midrule

OVERALL
& Score $\downarrow$
& \bestresult{12.42 $\pm$ 0.30}
& \secondresult{12.69 $\pm$ 0.40}
& 12.93 $\pm$ 0.22
& 12.83 $\pm$ 0.15
& 13.11 $\pm$ 0.38
& 13.13 $\pm$ 0.54
& 12.74 $\pm$ 0.63 \\

\bottomrule
\end{tabular}

\caption{Modality and component ablations on MIX-battery. Struct.
Proxy denotes the endpoint-derived structural descriptor. 
$dQ/dV$ denotes the cell-level lifecycle terrain. NDC, SOH,
CF, NCC, and Overall are scaled by \(10^2\), and CFR by \(10^4\).
Lower is better. Best results are shaded and bold; second-best results
are bold. Highlighting is determined using the unrounded mean values.}
\label{tab:complete_ablation}
\end{table*}

\begin{table*}[t]
\centering
\small
\begin{tabular}{llc|cc|ccc}
\toprule
& & \multirow{2}{*}{\textbf{Full STEK}}
& \multicolumn{2}{c|}{\textbf{Fusion Controls}}
& \multicolumn{3}{c}{\textbf{Koopman Controls}} \\
\cmidrule(lr){4-5}
\cmidrule(lr){6-8}

\textbf{Task} & \textbf{Metric} &
& \shortstack{\textbf{Direct Concat}}
& \shortstack{\textbf{FiLM w/o Quality}}
& \textbf{Shared K.}
& \shortstack{\textbf{State-Cond. K.}}
& \shortstack{\textbf{w/o K. Reg.}} \\
\midrule

\multirow{2}{*}{NDC}
& MAE $\downarrow$
& \bestresult{0.90 $\pm$ 0.06}
& 1.17 $\pm$ 0.21
& 0.98 $\pm$ 0.12
& \secondresult{0.94 $\pm$ 0.06}
& 0.94 $\pm$ 0.09
& 0.94 $\pm$ 0.07 \\

& MaxAE $\downarrow$
& \secondresult{7.65 $\pm$ 1.15}
& 10.02 $\pm$ 0.54
& 7.80 $\pm$ 0.56
& 8.96 $\pm$ 2.43
& \bestresult{7.41 $\pm$ 0.44}
& 7.97 $\pm$ 2.11 \\

\midrule
\multirow{2}{*}{SOH}
& MAE $\downarrow$
& \secondresult{0.40 $\pm$ 0.03}
& 0.47 $\pm$ 0.04
& 0.42 $\pm$ 0.05
& \bestresult{0.38 $\pm$ 0.02}
& \secondresult{0.40 $\pm$ 0.03}
& 0.40 $\pm$ 0.04 \\

& RMSE $\downarrow$
& \secondresult{0.99 $\pm$ 0.02}
& 1.30 $\pm$ 0.16
& \secondresult{0.99 $\pm$ 0.01}
& 0.99 $\pm$ 0.03
& \bestresult{0.97 $\pm$ 0.06}
& 1.03 $\pm$ 0.13 \\

\midrule
\multirow{2}{*}{CF}
& MAE $\downarrow$
& \bestresult{0.38 $\pm$ 0.02}
& 0.48 $\pm$ 0.04
& 0.42 $\pm$ 0.03
& \bestresult{0.38 $\pm$ 0.02}
& \secondresult{0.40 $\pm$ 0.02}
& 0.40 $\pm$ 0.04 \\

& RMSE $\downarrow$
& \bestresult{0.97 $\pm$ 0.02}
& 1.27 $\pm$ 0.12
& \secondresult{0.97 $\pm$ 0.03}
& 0.99 $\pm$ 0.03
& \secondresult{0.97 $\pm$ 0.05}
& 1.00 $\pm$ 0.10 \\

\midrule
\multirow{2}{*}{CFR}
& MAE $\downarrow$
& \bestresult{1.33 $\pm$ 0.00}
& 1.49 $\pm$ 0.06
& 1.37 $\pm$ 0.04
& \secondresult{1.33 $\pm$ 0.01}
& 1.36 $\pm$ 0.03
& 1.37 $\pm$ 0.05 \\

& RMSE $\downarrow$
& \bestresult{3.00 $\pm$ 0.05}
& 3.24 $\pm$ 0.05
& 3.05 $\pm$ 0.07
& \secondresult{3.05 $\pm$ 0.02}
& 3.07 $\pm$ 0.07
& 3.07 $\pm$ 0.09 \\

\midrule
\multirow{2}{*}{ICR}
& MAE ($\mathrm{m}\Omega$) $\downarrow$
& \secondresult{6.54 $\pm$ 0.98}
& 7.73 $\pm$ 0.99
& 7.07 $\pm$ 0.92
& 7.52 $\pm$ 0.81
& 6.96 $\pm$ 0.90
& \bestresult{5.96 $\pm$ 0.31} \\

& RE (\%) $\downarrow$
& \secondresult{11.99 $\pm$ 2.45}
& 16.05 $\pm$ 4.41
& 13.04 $\pm$ 3.61
& 14.82 $\pm$ 0.78
& 12.32 $\pm$ 1.19
& \bestresult{11.57 $\pm$ 1.92} \\

\midrule
\multirow{2}{*}{NCC}
& MAE $\downarrow$
& \bestresult{0.85 $\pm$ 0.03}
& 1.16 $\pm$ 0.21
& 0.94 $\pm$ 0.08
& 0.98 $\pm$ 0.05
& \secondresult{0.89 $\pm$ 0.10}
& 0.92 $\pm$ 0.06 \\

& RMSE $\downarrow$
& \bestresult{1.31 $\pm$ 0.09}
& 1.92 $\pm$ 0.41
& 1.36 $\pm$ 0.09
& 1.45 $\pm$ 0.15
& \secondresult{1.34 $\pm$ 0.15}
& 1.40 $\pm$ 0.18 \\

\midrule
QTY
& RQE (\%) $\downarrow$
& \bestresult{6.74 $\pm$ 0.64}
& 18.84 $\pm$ 6.44
& 7.38 $\pm$ 2.44
& 10.80 $\pm$ 3.05
& \secondresult{7.24 $\pm$ 1.12}
& 9.58 $\pm$ 2.64 \\

\midrule
OVERALL
& Score $\downarrow$
& \bestresult{12.42 $\pm$ 0.30}
& 15.22 $\pm$ 0.75
& 13.10 $\pm$ 0.39
& 12.71 $\pm$ 0.34
& \secondresult{12.71 $\pm$ 0.28}
& 12.74 $\pm$ 0.63 \\

\bottomrule
\end{tabular}
\caption{Fusion and Koopman-inspired controls on MIX-battery.
Fusion controls use the same auxiliary inputs as Full STEK.
Shared K. uses a shared identity-centered lifted map, whereas
State-Cond. K. conditions it on the endpoint representation.
Lower values are better for all metrics. NDC, SOH, CF, NCC, and Overall
are scaled by \(10^2\), and CFR by \(10^4\). Best results are
shaded and bold; second-best results are bold.
Highlighting is determined using the unrounded mean values.}
\label{tab:targeted_design_analysis}
\end{table*}
\begin{figure}[t]
    \centering
    \includegraphics[width=\linewidth]{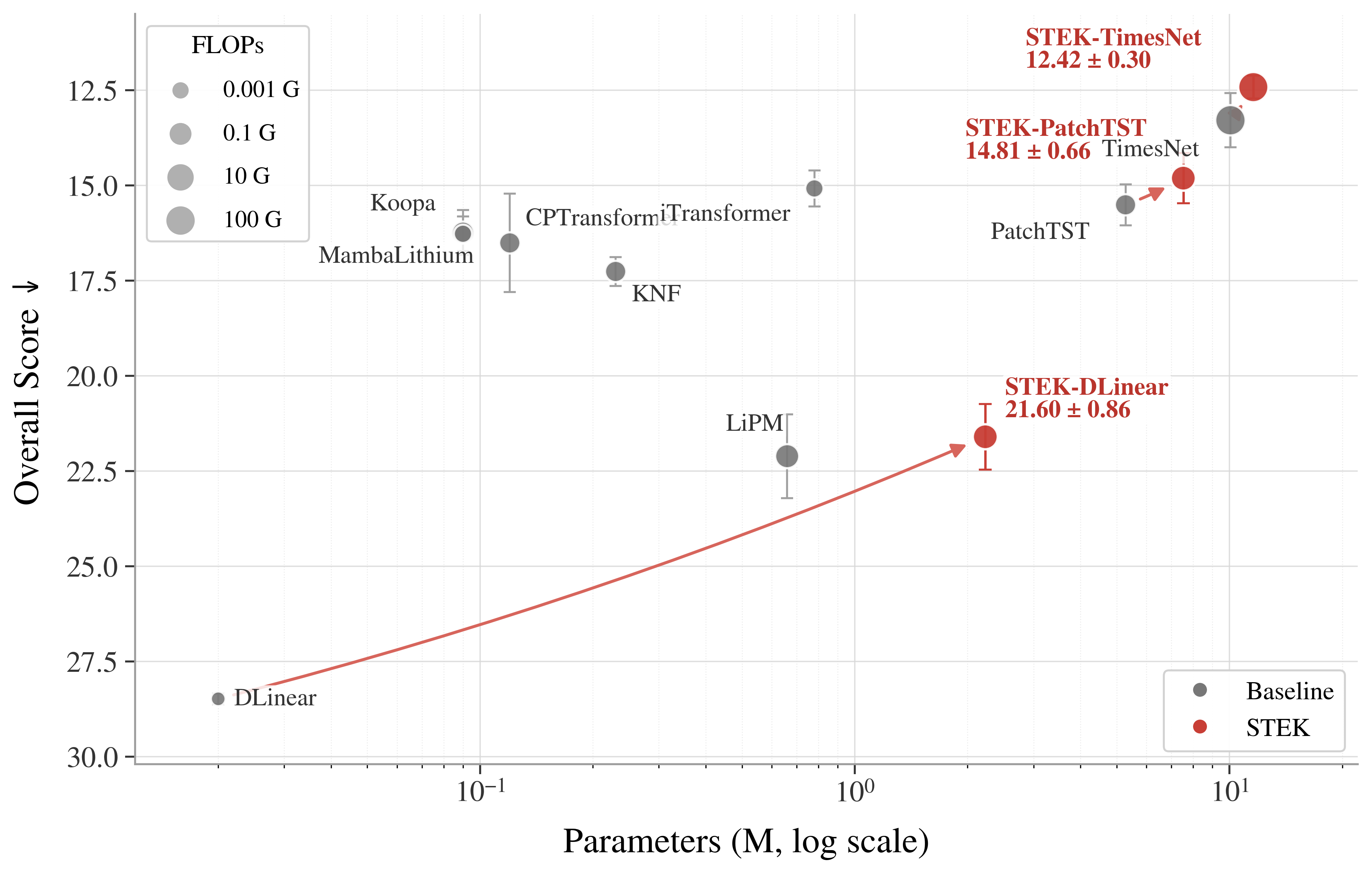}
    \caption{Performance--efficiency comparison under the fixed
MIX-battery seven-output setting. Bubble size represents FLOPs, and
lower Overall scores indicate better performance.}
\label{fig:efficiency}
\end{figure}

\begin{table*}[t]
\centering
\setlength{\tabcolsep}{3.5pt}
\begin{tabular}{llcccccc}
\toprule
& & \multicolumn{2}{c}{\textbf{CALCE}}
& \multicolumn{2}{c}{\textbf{HNEI}}
& \multicolumn{2}{c}{\textbf{MATR}} \\
\cmidrule(lr){3-4}
\cmidrule(lr){5-6}
\cmidrule(lr){7-8}

\textbf{Task} & \textbf{Metric}
& \textbf{Curve} & \textbf{Full}
& \textbf{Curve} & \textbf{Full}
& \textbf{Curve} & \textbf{Full} \\
\midrule

\multirow{2}{*}{NDC}
& MAE $\downarrow$
& 8.93 $\pm$ 0.74
& \textbf{8.39 $\pm$ 0.08}
& 2.93 $\pm$ 2.17
& \textbf{1.90 $\pm$ 0.43}
& 29.17 $\pm$ 8.95
& \textbf{13.55 $\pm$ 4.64} \\

& MaxAE $\downarrow$
& 14.42 $\pm$ 1.06
& \textbf{14.19 $\pm$ 0.61}
& \textbf{7.25 $\pm$ 0.51}
& 8.66 $\pm$ 3.66
& 43.43 $\pm$ 2.46
& \textbf{35.67 $\pm$ 2.47} \\

\multirow{2}{*}{SOH}
& MAE $\downarrow$
& 2.36 $\pm$ 0.77
& \textbf{1.86 $\pm$ 0.46}
& \textbf{1.59 $\pm$ 0.06}
& 1.60 $\pm$ 0.37
& 4.23 $\pm$ 0.62
& \textbf{2.75 $\pm$ 0.14} \\

& RMSE $\downarrow$
& 3.07 $\pm$ 0.83
& \textbf{2.74 $\pm$ 0.66}
& \textbf{1.72 $\pm$ 0.08}
& 1.78 $\pm$ 0.30
& 4.53 $\pm$ 0.50
& \textbf{3.29 $\pm$ 0.18} \\

\multirow{2}{*}{CF}
& MAE $\downarrow$
& 2.31 $\pm$ 0.76
& \textbf{1.86 $\pm$ 0.39}
& 1.28 $\pm$ 0.09
& \textbf{1.23 $\pm$ 0.39}
& 4.27 $\pm$ 0.67
& \textbf{2.44 $\pm$ 0.35} \\

& RMSE $\downarrow$
& 3.07 $\pm$ 0.83
& \textbf{2.74 $\pm$ 0.66}
& \textbf{1.72 $\pm$ 0.08}
& 1.78 $\pm$ 0.30
& 4.53 $\pm$ 0.50
& \textbf{3.29 $\pm$ 0.18} \\

\multirow{2}{*}{CFR}
& MAE $\downarrow$
& 12.11 $\pm$ 1.27
& \textbf{11.37 $\pm$ 0.19}
& 7.94 $\pm$ 0.48
& \textbf{7.13 $\pm$ 0.78}
& 2.90 $\pm$ 0.23
& \textbf{2.14 $\pm$ 0.36} \\

& RMSE $\downarrow$
& 14.59 $\pm$ 1.40
& \textbf{13.76 $\pm$ 0.49}
& 9.64 $\pm$ 0.51
& \textbf{8.78 $\pm$ 0.86}
& 3.11 $\pm$ 0.17
& \textbf{2.60 $\pm$ 0.28} \\

\multirow{2}{*}{ICR}
& MAE ($\mathrm{m}\Omega$) $\downarrow$
& 141.66 $\pm$ 15.84
& \textbf{126.18 $\pm$ 5.09}
& 48.17 $\pm$ 29.25
& \textbf{31.67 $\pm$ 11.34}
& \textbf{15.12 $\pm$ 7.20}
& 27.38 $\pm$ 16.18 \\

& RE (\%) $\downarrow$
& 79.16 $\pm$ 8.91
& \textbf{70.47 $\pm$ 2.80}
& 51.70 $\pm$ 31.41
& \textbf{34.44 $\pm$ 10.88}
& \textbf{46.70 $\pm$ 22.29}
& 84.65 $\pm$ 50.24 \\

\multirow{2}{*}{NCC}
& MAE $\downarrow$
& 6.70 $\pm$ 0.22
& \textbf{6.58 $\pm$ 0.20}
& 4.38 $\pm$ 2.24
& \textbf{2.16 $\pm$ 0.28}
& 28.66 $\pm$ 9.14
& \textbf{13.56 $\pm$ 4.02} \\

& RMSE $\downarrow$
& 7.25 $\pm$ 0.26
& \textbf{7.12 $\pm$ 0.12}
& 5.16 $\pm$ 2.17
& \textbf{2.80 $\pm$ 0.36}
& 31.06 $\pm$ 8.12
& \textbf{16.02 $\pm$ 4.60} \\

QTY
& RQE (\%) $\downarrow$
& 119.32 $\pm$ 69.50
& \textbf{82.44 $\pm$ 33.13}
& 89.42 $\pm$ 73.55
& \textbf{57.60 $\pm$ 58.05}
& 1310.13 $\pm$ 226.00
& \textbf{636.15 $\pm$ 339.97} \\

\midrule
OVERALL
& Score $\downarrow$
& 117.31 $\pm$ 17.44
& \textbf{105.33 $\pm$ 3.90}
& 67.93 $\pm$ 12.88
& \textbf{57.36 $\pm$ 8.33}
& 126.50 $\pm$ 25.10
& \textbf{70.50 $\pm$ 12.23} \\
\bottomrule
\end{tabular}
\caption{Holdout evaluation results for TimesNet under the curve-only
and full retrospective STEK settings. NDC, SOH, CF, NCC, and Overall are scaled by \(10^2\), while CFR metrics are scaled by $10^4$. Results are reported as mean $\pm$ standard deviation over three random seeds. Lower values are better, and the better result under each dataset is highlighted in bold.}
\label{tab:holdout_timesnet}
\end{table*}

\subsection{Experimental Setup}
We evaluate seven predefined cycle-level outputs. NDC and NCC
(SOC and CA in the implementation) are normalized capacity summaries
directly recoverable from a valid endpoint capacity channel. ICR and
QTY are same-cycle engineered quantities, SOH and CFR require historical
references, and CF is derived from SOH. Overall is computed as
$\mathrm{Overall}=\frac{1}{7}\sum_{j=1}^{7} \mathrm{MAE}_j/\max(\sigma_j^{\mathrm{train}},10^{-8})$
after restoring predictions to their original units, and is reported
on the $\times10^2$ scale. The seven-output interface is retained for
benchmark compatibility rather than treated as seven independent latent
states. Label-free rules achieve NDC/NCC MAEs of $0.0276/0.0392$ on the
$\times10^2$ scale; further details are provided in the supplementary
material.

Models are trained for up to 220 epochs with batch size 4, learning rate
$5\times10^{-5}$, dropout 0.2, $d_{\mathrm{model}}=64$, lifted dimension
96, and rank 8. Results are reported over seeds 2024--2026 using
validation-based checkpoint selection. Additional implementation details
are provided in the supplementary material.

\subsection{Overall Performance}
Endpoint-only baselines share the split, curve inputs, objective, and
checkpoint protocol, whereas STEK additionally uses archived lifecycle
context. Table~\ref{tab:mix_battery_comparison} therefore compares complete
settings, while Table~\ref{tab:targeted_design_analysis} provides
matched-input controls. STEK--TimesNet achieves the lowest Overall score,
$12.42\pm0.30$, reducing the endpoint-only iTransformer result by $17.6\%$.
This gain reflects both the additional archived context and the STEK design.

Under matched inputs, STEK reduces Overall by $18.4\%$ over direct
concatenation and by $5.2\%$ over FiLM without continuous quality weighting.
It also achieves the lowest learned-model errors on SOH, CF, ICR, and QTY,
and the second-lowest error on CFR, indicating a stronger aggregate
trade-off rather than uniform target-wise dominance.

\subsection{Generality Across Temporal Backbones}
Figure~\ref{fig:backbone_generality} shows Overall reductions of
\(6.5\%\), \(24.2\%\), and \(4.5\%\) for TimesNet, DLinear, and
PatchTST, respectively. All seven primary metrics improve for DLinear and PatchTST;
for TimesNet, six primary metrics improve while ICR MAE
degrades. The gains transfer
across the evaluated backbones, although learning remains unnecessary
for directly recoverable NDC and NCC.

\subsection{Ablation Study}
Table~\ref{tab:complete_ablation} separates modality and module
effects. Curve+Material+\(dQ/dV\) is the best reduced-modality setting
at \(12.69\), versus \(12.42\) for the full model. Removing FiLM or
the structural head raises Overall to \(13.11\) and \(13.13\).
Removing cross-channel regularization changes Overall from
\(12.42\pm0.30\) to \(12.74\pm0.63\) while improving ICR, indicating
a modest aggregate rather than uniform target-wise benefit.

QTY is the most variable target, with RQE from \(5.92\%\) to
\(10.81\%\). Because it depends on raw-cycle throughput but the input
lacks an explicit time channel, individual differences are not
interpreted further.

\subsection{Targeted Design Analysis}
Under matched inputs, Table~\ref{tab:targeted_design_analysis} shows
that Full STEK reduces Overall from \(15.22\) for direct concatenation
and \(13.10\) for quality-unaware FiLM to \(12.42\), improvements of
\(18.4\%\) and \(5.2\%\). It also lowers mean Overall by
\(2.3\%\)--\(2.5\%\) relative to the shared-map,
state-conditioned-map, and no-regularization variants. Given three
seeds and overlapping variability, this is supporting rather than
conclusive evidence for the lifted regularizer. The no-regularization
variant performs better on ICR, precluding uniform target-wise
superiority.

\begin{table*}[t]
\centering
\small
\setlength{\tabcolsep}{3.8pt}
\begin{tabular}{lcccccccc}
\toprule
\textbf{Descriptor}
& \multicolumn{6}{c}{\textbf{MAE} $\downarrow$}
& \textbf{RQE (\%)} $\downarrow$
& \textbf{Overall} $\downarrow$ \\
\cmidrule(lr){2-7}
\textbf{missing rate}
& \textbf{NDC}
& \textbf{SOH}
& \textbf{CF}
& \textbf{CFR}
& \textbf{ICR ($\mathrm{m}\Omega$)}
& \textbf{NCC}
& \textbf{QTY}
& \textbf{Score} \\
\midrule

0\%
& 0.90 $\pm$ 0.06
& 0.40 $\pm$ 0.03
& 0.38 $\pm$ 0.02
& 1.33 $\pm$ 0.00
& 6.54 $\pm$ 0.98
& 0.85 $\pm$ 0.03
& 6.74 $\pm$ 0.64
& 12.42 $\pm$ 0.30 \\

25\%
& 0.91 $\pm$ 0.05
& 0.40 $\pm$ 0.03
& 0.39 $\pm$ 0.01
& 1.33 $\pm$ 0.00
& 6.86 $\pm$ 0.91
& 0.87 $\pm$ 0.03
& 8.14 $\pm$ 0.90
& 12.52 $\pm$ 0.28 \\

50\%
& 0.92 $\pm$ 0.06
& 0.40 $\pm$ 0.03
& 0.38 $\pm$ 0.01
& 1.33 $\pm$ 0.00
& 7.09 $\pm$ 0.92
& 0.88 $\pm$ 0.05
& 9.15 $\pm$ 1.20
& 12.57 $\pm$ 0.28 \\

75\%
& 0.91 $\pm$ 0.09
& 0.40 $\pm$ 0.03
& 0.39 $\pm$ 0.02
& 1.33 $\pm$ 0.01
& 7.37 $\pm$ 0.97
& 0.88 $\pm$ 0.07
& 10.76 $\pm$ 1.73
& 12.65 $\pm$ 0.31 \\

100\%
& 0.90 $\pm$ 0.12
& 0.40 $\pm$ 0.03
& 0.39 $\pm$ 0.02
& 1.33 $\pm$ 0.01
& 7.63 $\pm$ 1.11
& 0.89 $\pm$ 0.10
& 12.15 $\pm$ 2.13
& 12.71 $\pm$ 0.36 \\

\bottomrule
\end{tabular}
\caption{Robustness to cell-level masking of the lifecycle terrain and
structural proxy. Intermediate rates use three model seeds and ten
masking seeds; 0\% and 100\% use three deterministic runs. NDC, SOH,
CF, NCC, and Overall are scaled by $10^2$, and CFR by $10^4$; ICR and
QTY are reported in m$\Omega$ and \%.}
\label{tab:modality_missingness}
\end{table*}

\subsection{Retrospective Cross-Source Transfer}
In three separate leave-one-source-out experiments, we hold
out CALCE, HNEI, or MATR from training, normalization,
validation, and model selection.
The full setting retains archived context for each
test cell. In Table~\ref{tab:holdout_timesnet}, it reduces Overall by
\(10.2\%\), \(15.6\%\), and \(44.3\%\), respectively. Gains are not
uniform: SOH changes little on HNEI, and ICR degrades on MATR for
reasons not isolated here. Archived context improves aggregate
retrospective transfer but does not eliminate source-specific
failures.

\subsection{Robustness to Descriptor Availability}
Table~\ref{tab:modality_missingness} independently masks the lifecycle
terrain and endpoint-derived structural proxy at the cell level.
Material context is excluded because it does not enter the inference
path. With both structural descriptors unavailable, Overall increases
from \(12.42\) to \(12.71\) (\(2.3\%\)); changes are concentrated in
ICR and QTY. This supports fallback when structural descriptors are absent.
This experiment does not evaluate robustness to noisy, corrupted, or misleading descriptors.

\subsection{Efficiency Study}
Figure~\ref{fig:efficiency} compares performance and inference cost.
Parameters include the training-only cross-channel branch; FLOPs cover
the prediction pass but exclude offline terrain construction and
precomputed material-text embeddings. STEK improves Overall across
all three backbones at the cost of parameters and FLOPs.

\section{Conclusion}
We presented STEK for retrospective harmonization of
cycle-level attributes in heterogeneous battery archives.
Because the lifecycle terrain may include cycles after endpoint
$t$, the method is not intended for online or prospective
prediction; when complete target-defining raw signals are
available, analytic computation remains preferable. Under the
fixed benchmark protocol, archived structural context with
quality-gated fusion improves aggregate learned-model
performance across three temporal backbones. Matched-input
controls attribute the larger gains to the fusion pathway, while
the training-only lifted regularizer provides a smaller aggregate
benefit and does not improve every target. Deterministic rules
substantially outperform learned models on NDC and NCC, so
the principal learned contribution concerns history-dependent
and cross-signal outputs under standardized inputs. Future work
will examine incomplete archives, corrupted modalities, and
prospective settings.

\bibliography{references}
\end{document}


\maketitle

\section{Dataset Details}

\subsection{Dataset Overview}

\begin{table*}[htbp]
\centering
\resizebox{\textwidth}{!}{%
\begin{tabular}{lllllrrrrr}
\toprule
\textbf{Source} & \textbf{Type} & \textbf{Format} &
\textbf{Cathode} & \textbf{Anode} & \textbf{Cap. (Ah)} &
\textbf{Temp. ($^\circ$C)} & \textbf{Cells} &
\textbf{Cycles} & \textbf{Prot.} \\
\midrule
CALCE & Li-ion & Prismatic & LCO & Graphite
& 1.10 & 25 & 13 & 14,299 & 2 \\
HNEI & Li-ion & 18650 & LCO+NCM442 & Graphite
& 2.80 & 25 & 14 & 15,164 & 2 \\
MATR & Li-ion & 18650 & LFP & Graphite
& 1.10 & 30 & 169 & 140,001 & 81 \\
RWTH & Li-ion & 18650 & NMC & Carbon
& 3.00 & 25 & 48 & 105,006 & 1 \\
MICH & Li-ion & Pouch & NCM111 & Graphite
& 2.36 & 25, 45 & 40 & 19,895 & 2 \\
MICH\_EXP & Li-ion & Pouch & NCM111 & Graphite
& 5.00 & $-5$, 25, 45 & 18 & 6,545 & 6 \\
HUST & Li-ion & 18650 & LFP & Graphite
& 1.10 & 30 & 77 & 146,120 & 77 \\
Stanford & Li-ion & 18650 & NMC532 & AG
& 0.24 & 30 & 41 & 39,329 & 1 \\
Na-ion & Na-ion & 18650 & Unknown & Unknown
& 1.00 & 25, 30 & 64 & 12,604 & 12 \\
CALB & Li-ion & Prismatic & NMC & Graphite
& 58.0 & 0, 25, 35, 45 & 27 & 2,791 & 2 \\
\bottomrule
\end{tabular}}
\caption{Source statistics for MIX-battery. Capacity denotes the
source-reported nominal capacity, which is retained as a dataset
descriptor; harmonized values used for curve normalization are
specified in the preprocessing section. AG denotes artificial
graphite, and Prot. denotes the number of cycling protocols.}
\label{tab:dataset_stats}
\end{table*}

MIX-battery contains 511 candidate cells and 501,754 cycles from ten
public sources. It covers Li-ion and Na-ion cells, multiple cathode
families and form factors, source-reported nominal capacities from
0.24 to 58.0\,Ah, temperatures from \(-5^\circ\)C to \(45^\circ\)C,
and diverse cycling protocols. This variation supports retrospective
evaluation across heterogeneous chemistries, scales, and operating
conditions.

The constituent datasets are CALCE~\cite{33},
HNEI~\cite{34}, MATR~\cite{19,37}, RWTH~\cite{39},
MICH~\cite{3}, MICH\_EXP~\cite{38}, HUST~\cite{35},
Stanford~\cite{41}, Na-ion~\cite{31}, and CALB~\cite{31}.
All sources are converted into the same canonical record format before
cell splitting and endpoint construction. Table~\ref{tab:dataset_stats}
summarizes their source-level characteristics; preprocessing,
eligibility, and retrospective lifecycle-context construction are
described in the following sections.

\section{Additional Reproducibility Details}
\label{app:reproducibility_details}

\paragraph{Baseline configurations.}
All endpoint-only neural baselines use the same resampled cycling-curve
input, cell split, seven-output objective, optimizer, training budget,
and validation-based checkpoint-selection protocol. They do not use
STEK's archived lifecycle terrain, endpoint-derived structural proxy,
or material-text context. STEK and the explicitly marked multimodal
fusion control additionally use these auxiliary descriptors.

The common training setting uses 100 resampled points per curve, batch
size 4 with 8-step gradient accumulation, learning rate
\(5\times10^{-5}\), dropout 0.2, and seeds 2024--2026. A hidden width
of 64 is used where compatible, while model-specific blocks follow
their corresponding reference implementations. All models are adapted
to the common seven-output prediction interface.

\begin{table}[ht]
\centering
\footnotesize
\setlength{\tabcolsep}{3pt}
\begin{tabular}{lccc}
\toprule
Model & Input & Width & Principal adaptation \\
\midrule
TimesNet & curves & 64 & seven-output prediction head \\
DLinear & curves & 64 & seven-output prediction head \\
PatchTST & curves & 64 & seven-output prediction head \\
iTransformer & curves & 64 & seven-output prediction head \\
MambaLithium & curves & 64 & canonical-curve adaptation \\
LiPM & curves & 64 & canonical-curve adaptation \\
CPTransformer & curves & 64 & common seven-output objective \\
Koopa & curves & 64 & 2 blocks; segment length 24 \\
KNF & curves & 64 & 3 MLP layers; local rank 8 \\
\bottomrule
\end{tabular}
\caption{Principal baseline settings used in the common seven-output
comparison. Native architectural components are retained where
compatible, while all models use the same prediction interface and
evaluation protocol.}
\label{tab:baseline_config}
\end{table}

\paragraph{\(dQ/dV\) lifecycle-terrain construction.}
For cell \(b\), an exported lifecycle terrain is constructed from the
charge curves in its archived cycling trajectory
\[
\mathcal{X}_b
=
\{X_{b,\tau}\}_{\tau=1}^{N_b}.
\]
The resulting cell-level terrain is shared across the evaluated
endpoints of that cell. This construction belongs to the retrospective
archived-context setting and is not restricted to cycles available by
a particular endpoint \(t\).

For each selected archived cycle, charge capacity \(Q\) and voltage
\(V\) are sorted by voltage and smoothed using a Savitzky--Golay filter
with window length 9 and polynomial order 3 when enough points are
available. Incremental capacity is then computed as
\[
\frac{\mathrm{d}Q}{\mathrm{d}V}
=
\frac{\nabla Q}{\nabla V+\epsilon},
\]
with protection against small voltage steps. Each cycle profile is
interpolated onto a per-cell voltage grid of 200 bins spanning the
finite voltage range of the selected archived cycles.

Cycle columns are selected using an adaptive stride and subsequently
resized to 64 columns. Missing or invalid columns inherit the previous
valid column when available and otherwise remain zero. Non-finite
values are replaced by zero. Visualization clips values at the 99th
percentile of absolute magnitude, whereas the stored model input
retains the float32 terrain matrix.

During loading, the terrain is resized to
\(1\times200\times64\) when necessary, transformed by
\(\log(1+|\cdot|)\), normalized by its maximum absolute value, and
then sample-normalized and clamped to \([-5,5]\) by the structural
encoder.

If no exported cell-level terrain is available, the loader constructs
a tensor-compatible fallback from the endpoint voltage, current, and
capacity curve. This fallback contains only endpoint-local information
and is not interpreted as an archived lifecycle terrain.

\paragraph{Material-text metadata representation.}
Material context is stored as a cell-aligned metadata representation.
Each cell receives a battery-type embedding and, when available, a
768-dimensional static material-text vector. Metadata are obtained
from file-name parsing and available cell- or source-level records for
cathode, anode, electrolyte, form factor, nominal capacity,
temperature, and C-rate.

Cathode, anode, and electrolyte labels are mapped to predefined
material strings. The dictionary includes LFP, LCO, NCM/NCA variants,
graphite, zinc, common carbonate solvents, and LiPF6. Multi-material
systems are represented as weighted mixtures, with cathode, anode, and
electrolyte weights of 0.35, 0.35, and 0.30, respectively. Known
mixtures such as LCO+NCM442 are expanded before weighting. Unknown
electrolytes use the placeholder
\texttt{Commercial\_Standard} or an optional unknown-electrolyte
token.

MolT5 encodes the resulting strings, and the component embeddings are
combined into one static vector per cell. These vectors represent
available material-text metadata rather than verified cell-specific
molecular structures. When cell-level chemistry is unavailable, some
fields inherit source-level metadata and may consequently encode
source-level context.

\paragraph{Multimodal concatenation control.}
The direct-concatenation control receives the same auxiliary
information as STEK: material-text context, the \(dQ/dV\) lifecycle
terrain, and the 64-dimensional endpoint-derived structural proxy.
The temporal encoder is unchanged. The terrain and structural proxy
are encoded as \(s\in\mathbb{R}^{64}\), while the battery-type and
material-text branch produces \(m\in\mathbb{R}^{64}\).

This control disables quality-gated FiLM and Koopman-inspired
regularization, concatenates \([s;m]\), projects it to 64 dimensions,
and combines it with the temporal representation before the prediction
head. It provides a matched-input fusion control, although exact
parameter matching is not imposed.

\paragraph{Definitions of fusion and Koopman controls.}
The \emph{FiLM w/o Quality} control, more precisely \emph{FiLM w/o
continuous quality weighting}, retains the material, lifecycle-terrain,
endpoint-derived structural-proxy, conditioned-Koopman, structural-head,
and residual-prediction pathways of Full STEK. It replaces the learned
continuous terrain and proxy quality scores with binary availability
indicators,
\begin{equation}
q_{T,b}\leftarrow v_{T,b},\qquad
q_{R,b,t}\leftarrow v_{R,b,t},
\end{equation}
and removes the continuous quality multiplier from the outer physical
gate, so that
\begin{equation}
g_{\mathrm{phy}}
= g_{\mathrm{warmup}}\,\mathbb{I}_{\mathrm{available}}.
\end{equation}
Binary missing-modality masking, fallback to the original endpoint
representation when structural inputs are unavailable, the internal
learned FiLM gate, training warm-up, and the learned validity-gated
residual blend remain active. Material context continues to condition
only the lifted operator rather than FiLM. The conditioned cross-channel
regularizer and its loss weight of \(0.05\) are unchanged.

The \emph{Shared K.} control replaces the context-conditioned factors
with globally learned parameters
\begin{equation}
U,V\in\mathbb{R}^{d_K\times r},\qquad
\delta\in\mathbb{R}^{d_K}.
\end{equation}
The operator retains the same bounded near-identity construction,

\begin{equation}
\begin{aligned}
A &= \frac{\tanh(U)\tanh(V)^{\top}}{\sqrt{r}} + \operatorname{diag}(\tanh\delta), \\
K &= \alpha I + \eta_K\frac{A}{\max(1,\lVert A\rVert_F)}.
\end{aligned}
\end{equation}

Here, the memory vector \(m_K\), scalar modulation \(d\), and
sample-independent operator-update scale \(\eta_K\) are also globally
learned. Material and structural context do not enter \(K\). FiLM, the
structural head, the validity-gated residual branch, auxiliary inputs,
and loss weights otherwise remain the same as in Full STEK. This control
is not exactly parameter-matched: the implementation retains inactive
hypernetwork parameters while adding the global operator parameters. At
\(d_K=96\) and \(r=8\), the directly introduced lifted-operator
parameters add 1,729 trainable parameters, and the full-model parameter
difference is slightly larger because an additional inactive operator
instance remains instantiated.

The \emph{State-Cond. K.} control retains the operator hypernetwork but
zeros its material and structural inputs. Its effective context is
\begin{equation}
c_{K,b,t}=\operatorname{LN}\!\left[
0\,\Vert\,0\,\Vert\,\operatorname{Pool}(\widetilde Z_{b,t})
\right].
\end{equation}
Consequently, \(U_{b,t}\), \(V_{b,t}\), \(\delta_{b,t}\),
\(m_{K,b,t}\), and \(d_{b,t}\) depend only on the pooled endpoint
representation. The structure gate controlling \(\eta_K\) receives an
all-zero structural vector and therefore reduces to a sample-independent
learned constant. The remaining input pathways, FiLM, structural head,
validity-gated residual branch, losses, and hyperparameters are unchanged.
Its parameter count matches the Full conditioned-operator model, although
some hypernetwork input connections always receive zero values.

All three controls use the same modalities, lift dimension, rank, split,
target objective, loss weights, training budget, seed set, and
validation-based checkpoint-selection rule as Full STEK, subject to the
parameter-matching caveat for Shared K described above.

\section{Data Processing and Sample Construction}

\subsection{Canonical Records and Cycle Inputs}

Each cell is stored as a canonical pickle record containing
\path{voltage_in_V}, \path{current_in_A},
\path{charge_capacity_in_Ah},
\path{discharge_capacity_in_Ah}, and
\path{time_in_s}, expressed in V, A, Ah, Ah, and s, respectively.
Unit conversion is completed during canonicalization. Negative
capacity entries are marked invalid and back-filled within each cycle.

The loader reads the normalization capacity
\(Q_b^{\mathrm{nom}}\) from
\texttt{nominal\_capacity\_in\_Ah}. For RWTH records, the fixed
harmonized metadata used by the benchmark preprocessing pipeline specify
\(Q_b^{\mathrm{nom}}=1.85\) Ah for curve and target normalization,
whereas the source-reported 3.0\,Ah value in
Table~\ref{tab:dataset_stats} is retained only as a dataset
descriptor. The canonicalization metadata, rather than the source-level
descriptor table, are treated as authoritative for model normalization.
This harmonized value is applied before splitting and is not selected or
inferred from validation or test performance; the discrepancy therefore
reflects benchmark harmonization rather than data-dependent tuning.

Charge and discharge segments are separated using a \(0.01\)C
threshold, resampled to 50 points each, and concatenated. The endpoint
input is
\begin{align}
\overline{\mathbf V}_{b,t}
&=
\frac{\mathbf V_{b,t}}{\max_k V_{b,t,k}},
\label{eq:voltage_normalization}\\
\overline{\mathbf I}_{b,t}
&=
\frac{\mathbf I_{b,t}}{Q_b^{\mathrm{nom}}},
\label{eq:current_normalization}\\
\overline{\mathbf Q}_{b,t}
&=
\frac{\mathbf Q_{b,t}}{Q_b^{\mathrm{nom}}},
\label{eq:capacity_normalization}\\
\mathbf X_{b,t}
&=
\begin{bmatrix}
\overline{\mathbf V}_{b,t}\\
\overline{\mathbf I}_{b,t}\\
\overline{\mathbf Q}_{b,t}
\end{bmatrix}
\in\mathbb{R}^{3\times100}.
\label{eq:cycle_input}
\end{align}
Thus, \(\mathbf X_{b,t}\) contains normalized endpoint voltage,
C-rate, and capacity.

Candidate endpoints satisfy
\begin{equation}
t
\in
\{96,\ldots,\min(100,\mathrm{EOL}_b-1)\}.
\label{eq:candidate_endpoints}
\end{equation}
Each example contains one endpoint curve \(\mathbf X_{b,t}\), and a
cell contributes at most five examples. Cycle 96 is the earliest
evaluated endpoint, not the number of input cycles.

\subsection{Definitions of Prediction Targets}

For cycle \(t\) of cell \(b\), define
\begin{equation}
D_{b,t}
=
\max_k Q^{\mathrm{dis}}_{b,t,k},
\qquad
C_{b,t}
=
\max_k Q^{\mathrm{chg}}_{b,t,k}.
\label{eq:cycle_capacities}
\end{equation}
The five capacity-related outputs are
\begin{align}
\mathrm{NDC}_{b,t}
&=
\frac{D_{b,t}}{Q_b^{\mathrm{nom}}},
\label{eq:ndc_target}\\
\mathrm{SOH}_{b,t}
&=
\frac{D_{b,t}}{D_{b,1}},
\label{eq:soh_target}\\
\mathrm{CF}_{b,t}
&=
1-\mathrm{SOH}_{b,t},
\label{eq:cf_target}\\
\mathrm{CFR}_{b,t}
&=
\frac{\mathrm{SOH}_{b,t-5}-\mathrm{SOH}_{b,t}}{5},
\label{eq:cfr_target}\\
\mathrm{NCC}_{b,t}
&=
\operatorname{clip}\!\left(
\frac{C_{b,t}}{Q_b^{\mathrm{nom}}},
0,2
\right).
\label{eq:ncc_target}
\end{align}

NDC and NCC correspond to the outputs denoted SOC and CA,
respectively, in the implementation and stored checkpoints. NDC is a
normalized discharge-capacity ratio rather than instantaneous state of
charge, while NCC is normalized charge capacity rather than a direct
electrochemical charge-acceptance measurement. Renaming these outputs
does not change their labels, training procedure, or evaluation. SOH
measures first-cycle capacity retention, CF is its complement, and CFR
is its five-cycle backward difference.

NDC and NCC are deterministic summaries of the underlying raw
capacity record. Learned-model performance on these two outputs is
therefore interpreted as joint curve-summary recovery under
resampling, missing segments, and recording heterogeneity rather than
as evidence of latent-state inference.

The DC-resistance proxy ICR is
\begin{equation}
\mathrm{ICR}_{b,t}
=
10^3\,
\operatorname{median}\!\left(
\left|
\frac{\Delta V}{\Delta I}
\right|
\right).
\label{eq:icr_target}
\end{equation}
Differences are evaluated around current-sign transitions within a
\(\pm3\)-sample window and clipped to \([0,500]\) m\(\Omega\).
A robust current-change estimate is used when no transition is valid.

The current-throughput proxy QTY uses
\begin{align}
\Delta\tau_{b,t,k}
&=
\max(\tau_{b,t,k+1}-\tau_{b,t,k},0),\\
A_{b,t}
&=
\frac{1}{3600}
\sum_k
|I_{b,t,k}|\,\Delta\tau_{b,t,k},
\end{align}
and
\begin{equation}
\mathrm{QTY}_{b,t}
=
\operatorname{clip}(A_{b,t},0,50).
\label{eq:qty_target}
\end{equation}
QTY is reported in Ah, and non-finite values are replaced by zero.

\subsection{Deterministic Capacity-Recovery Diagnostic}

Because NDC and NCC are deterministic summaries of the endpoint
capacity record, we evaluate direct input-derived estimators. Given
the normalized charge and discharge segments
\(\overline{\mathbf Q}_{b,t}^{\mathrm{chg}}\) and
\(\overline{\mathbf Q}_{b,t}^{\mathrm{dis}}\), the segment-maximum
rules are
\begin{equation}
\begin{aligned}
\widehat{\mathrm{NDC}}_{b,t}^{\mathrm{seg}}
&=
\max_k \overline Q_{b,t,k}^{\mathrm{dis}},\\
\widehat{\mathrm{NCC}}_{b,t}^{\mathrm{seg}}
&=
\operatorname{clip}\!\left(
\max_k \overline Q_{b,t,k}^{\mathrm{chg}},
0,2
\right).
\end{aligned}
\label{eq:deterministic_capacity_rules}
\end{equation}

Four test endpoints from one CALB cell contain an identically zero
resampled discharge-capacity segment. We therefore also report the
following label-free missing-segment safeguard:
\begin{equation}
\widehat{\mathrm{NDC}}_{b,t}^{\mathrm{fallback}}
=
\begin{cases}
\max_k \overline Q_{b,t,k}^{\mathrm{dis}},
& \max_k \overline Q_{b,t,k}^{\mathrm{dis}}>0,\\
\max_k \overline Q_{b,t,k}^{\mathrm{chg}},
& \text{otherwise}.
\end{cases}
\label{eq:ndc_zero_segment_fallback}
\end{equation}
The safeguard uses neither target labels nor lifecycle or material
context at inference. Because it addresses a missing-segment failure,
we report it as an input-recoverability diagnostic rather than as a
seven-output model-selection baseline. NCC is unchanged because the
corresponding charge segments are available.

\begin{table}[htbp]
\centering
\setlength{\tabcolsep}{3.5pt}
\resizebox{\columnwidth}{!}{%
\begin{tabular}{lcccc}
\toprule
\textbf{Method}
& \textbf{NDC MAE}
& \textbf{NDC MaxAE}
& \textbf{NCC MAE}
& \textbf{NCC RMSE} \\
\midrule
STEK--TimesNet
& 0.9022
& 7.6520
& 0.8507
& 1.3079 \\
Segment-maximum rule
& 1.0121
& 102.2336
& 0.0392
& 0.1384 \\
Segment maximum + zero-segment fallback
& 0.0276
& 0.8163
& 0.0392
& 0.1384 \\
\bottomrule
\end{tabular}}
\caption{Input-derived capacity-recovery diagnostic on the
MIX-battery test split. All errors are multiplied by \(10^2\).
The deterministic rules require neither training nor target-label
access. The plain NDC rule is affected by four endpoints with an
identically zero resampled discharge segment; over the remaining
410 endpoints, its NDC MAE and MaxAE are 0.0247 and 0.8163,
respectively. The STEK values are included for scale, with seed-level
uncertainty reported in the main comparison. Rule-based rows apply
only to NDC and NCC and therefore do not receive an Overall score.}
\label{tab:deterministic_capacity_rules}
\end{table}

The deterministic rules substantially outperform the learned model
on NDC and NCC once the missing discharge segment is handled. These
results confirm that the two quantities are directly recoverable from
a valid endpoint capacity channel. Their inclusion preserves
compatibility with the fixed seven-output benchmark and tests whether
a unified model can reproduce simple capacity summaries alongside
more demanding outputs; their errors are not treated as evidence of
latent-state inference.

For ICR and QTY, we distinguish two types of analytic reference.
A definition-reproduction oracle could use the original voltage,
current, and time records to rerun the label-generation formulas.
Such an oracle verifies label and implementation consistency but is
not an input-matched model baseline. In contrast, an input-matched
analytic baseline may use only the resampled
\(\mathbf X_{b,t}\). Exact reconstruction is unavailable under this
restriction: QTY requires timestamps omitted from
\(\mathbf X_{b,t}\), whereas ICR depends on absolute voltage and
current scales altered by input normalization. We therefore do not
present definition-reproduction oracle values as competing model
results or claim exact input-matched recovery for ICR or QTY.

\begin{table}[htbp]
\centering
\setlength{\tabcolsep}{3.5pt}
\begin{tabular}{@{}p{0.28\columnwidth}p{0.18\columnwidth}
p{0.43\columnwidth}@{}}
\toprule
\textbf{Category} & \textbf{Outputs} &
\textbf{Required information} \\
\midrule
Directly recoverable & NDC, NCC & Endpoint capacity channel \\
Same-cycle engineered & ICR, QTY &
Raw cross-signal or time record \\
History-dependent & SOH, CFR & First or preceding cycles \\
Algebraically derived & CF & \(1-\mathrm{SOH}\) \\
\bottomrule
\end{tabular}
\caption{Information requirements of the seven benchmark outputs.
The categories describe target construction rather than statistical
independence.}
\label{tab:target_information_groups}
\end{table}

\paragraph{Task scope.}
The benchmark evaluates retrospective cycle-level multi-target
estimation. Every setting uses an endpoint-local charge--discharge
curve. The terrain-enabled STEK setting additionally uses a cell-level
lifecycle terrain constructed from the archived trajectory, whereas
endpoint-only baselines do not receive this lifecycle context.

The outputs span directly recoverable curve summaries, same-cycle
engineered quantities, history-dependent quantities, and an
algebraically derived output, as summarized in
Table~\ref{tab:target_information_groups}. The principal learning
problem concerns the joint estimation of history-dependent and
cross-signal quantities, while NDC and NCC serve as recoverability
controls within the fixed output interface. CF is retained for
compatibility with that interface but is not an independent target
because it is exactly \(1-\mathrm{SOH}\).

\subsection{Eligibility and Label Availability}

The predefined cell-level splits contain 326, 92, and 93 candidate
training, validation, and test cells. Endpoint filtering retains 302,
86, and 84 cells, contributing 1,501, 426, and 414 examples,
respectively. The final evaluation therefore uses 472 cells and 2,341
examples. All endpoints from a cell remain in the same split, and all
preprocessing statistics and label scalers are fitted exclusively on
the 1,501 training examples.

\subsection{Construction of the Endpoint-Derived Structural Proxy}
\label{app:structural_proxy}

The implementation historically refers to this 64-dimensional
descriptor as the relaxation proxy. More precisely, it is an
endpoint-derived structural summary of voltage, current, capacity,
their first differences, and incremental-capacity shape. It does not
estimate a physical relaxation time constant. We therefore refer to it
below as the \emph{endpoint-derived structural proxy}.

Let \(\bar V_{b,t}\), \(\bar I_{b,t}\), and \(\bar Q_{b,t}\) denote
the valid voltage, current, and capacity sequences of endpoint \(t\).
Their first differences and stabilized incremental-capacity signal are
\begin{equation}
\begin{aligned}
\Delta V_{b,t,j}
&=
\bar V_{b,t,j+1}-\bar V_{b,t,j},\\
\Delta Q_{b,t,j}
&=
\bar Q_{b,t,j+1}-\bar Q_{b,t,j},\\
g_{b,t,j}
&=
\operatorname{clip}\!\left(
\frac{\Delta Q_{b,t,j}}
{|\Delta V_{b,t,j}|+\epsilon_g},
-c_g,c_g
\right),
\end{aligned}
\label{eq:proxy_signals}
\end{equation}
where \(\epsilon_g=10^{-6}\) and \(c_g=50\).

For a sequence \(u=(u_1,\ldots,u_n)\), define

\begin{equation}
\begin{aligned}
\phi(u) = [&
\mu(u),\sigma(u),\min(u),\max(u),\\
&u_1,u_n,\operatorname{range}(u),\operatorname{rms}(u)
],
\end{aligned}
\label{eq:proxy_statistics}
\end{equation}

where
\begin{equation}
\operatorname{range}(u)
=
\max(u)-\min(u),
\qquad
\operatorname{rms}(u)
=
\sqrt{\frac{1}{n}\sum_{j=1}^{n}u_j^2}.
\end{equation}

The final descriptor is
\begin{equation}
\begin{aligned}
\mathbf r_{b,t}
={}&[
\phi(\bar V_{b,t})
\Vert\phi(\bar I_{b,t})
\Vert\phi(\bar Q_{b,t})
\Vert\phi(\Delta V_{b,t})\\
&\Vert\phi(\Delta Q_{b,t})
\Vert\phi(g_{b,t})
\Vert\mathcal R_{16}(g_{b,t})
],
\end{aligned}
\label{eq:structural_proxy}
\end{equation}
where \(\mathcal R_{16}\) linearly resamples \(g_{b,t}\) to 16
points. The six statistical groups contribute \(6\times8=48\)
features, and the resampled profile contributes 16. All features use
only the endpoint curve, and their normalization is fitted using
training cells only.

\section{Training and Evaluation Details}
\label{app:training_evaluation}

\subsection{Implementation Settings}

Candidate endpoints are cycles 96--100. Each example contains one
endpoint charge--discharge curve resampled to 100 points, while the
full STEK setting additionally uses the archived cell-level lifecycle
terrain and endpoint-derived structural proxy. Normalization statistics
are fitted only on training cells. Validation cells are used for early
stopping and checkpoint selection, and test cells are used only for
final evaluation.

\begin{table}[t]
\centering
\begin{tabular}{lc}
\toprule
\textbf{Hyperparameter} & \textbf{Value} \\
\midrule
Optimizer & AdamW \\
Learning rate / weight decay & $5\times10^{-5}$ / $10^{-4}$ \\
Maximum epochs & 220 \\
Batch size / accumulation & 4 / 8 \\
Dropout & 0.2 \\
Hidden dimension / default layers & 64 / 2 where applicable \\
Resampled points per endpoint & 100 \\
Random seeds & 2024, 2025, 2026 \\
\bottomrule
\end{tabular}
\caption{Default common training settings. Architecture-specific depths
follow Table~\ref{tab:baseline_config} and the corresponding reference
implementations.}
\label{tab:implementation_settings}
\end{table}

\begin{table}[t]
\centering
\begin{tabular}{@{}>{\raggedright\arraybackslash}p{0.56\columnwidth}
>{\centering\arraybackslash}p{0.30\columnwidth}@{}}
\toprule
\textbf{Component} & \textbf{Setting} \\
\midrule
Identity coefficient \(\alpha\) & 0.90 \\
Base operator-update scale \(\eta_0\) & 0.05 \\
Conditioned update scale \(\eta_{K,b,t}\)
& \((0.0125,0.05)\) \\
Frobenius perturbation scaling
& \(A/\max(1,\lVert A\rVert_F)\) \\
Lifted dimension / rank & 96 / 8 \\
Scalar modulation range \(d_{b,t}\)
& \((0.90,1.10)\) \\
Mixing coefficient scale & 0.05 \\
Nonlinear residual scale \(\eta_r\) & 0.05 \\
FiLM scale \(\eta_s\) & 0.05 \\
FiLM bounds \((\delta_\gamma,\delta_\beta)\) & (0.10, 0.10) \\
Residual-gate bound \(\lambda_{\max}\) & 0.16 \\
Loss weight \(\lambda_{\mathrm{cc}}\) & 0.05 \\
\(dQ/dV\) terrain size & \(1\times200\times64\) \\
Endpoint structural-proxy dimension & 64 \\
Material-text encoder & Frozen MolT5-base \\
Material-text projection & \(768\rightarrow128\) \\
Structural / material dimensions & 128 / 128 \\
\bottomrule
\end{tabular}
\caption{Default full-model STEK settings. The conditioned lifted
linear map and its consistency loss form the training-only
Koopman-inspired regularization branch.}
\label{tab:stek_detailed_settings}
\end{table}

All methods use the same validation-based checkpoint-selection rule.

\subsection{Prediction and Koopman-Inspired Cross-Channel
Regularization}

For endpoint \(t\) of cell \(b\), the loader returns
\(X_{b,t}\in\mathbb{R}^{3\times100}\), whose channels are voltage,
current, and capacity. The backbone output is
\begin{equation}
Z_{b,t}
=
[z_{b,t}^{V},z_{b,t}^{I},z_{b,t}^{Q}]
\in\mathbb{R}^{3\times d}.
\end{equation}
Structural FiLM modulates these representations, and the residual head
provides the inference-time correction
\begin{equation}
\hat{\mathbf{y}}_{b,t}
=
\hat{\mathbf{y}}_{b,t}^{\mathrm{base}}
+
\lambda_{b,t}\odot
\Delta\hat{\mathbf{y}}_{b,t}^{\mathrm{struct}}.
\label{eq:appendix_prediction_path}
\end{equation}

The Koopman-inspired branch supplies only a training-time
cross-channel constraint. Each modulated channel is lifted as
\[
\psi_{b,t}^{c}
=
\mathcal{E}_{\mathrm{lift}}(\widetilde Z_{b,t}^{c}),
\qquad c\in\{V,I,Q\}.
\]
The operator context concatenates material, structural, and pooled
curve representations and applies layer normalization:
\begin{equation}
c_{K,b,t}
=
\operatorname{LN}\!\left(
[m_b\Vert s_{b,t}\Vert
\operatorname{Pool}(\tilde Z_{b,t})]
\right).
\label{eq:appendix_koopman_context}
\end{equation}
Three operator heads generate raw low-rank factors and a diagonal
component. The bounded perturbation and sample-conditioned update are

\begin{equation}
\begin{aligned}
U_{b,t}^{\mathrm{raw}} &= \Phi_U(c_{K,b,t}), \quad
V_{b,t}^{\mathrm{raw}} = \Phi_V(c_{K,b,t}), \\
\delta_{b,t}^{\mathrm{raw}} &= \Phi_D(c_{K,b,t}), \\
A_{b,t} &= \frac{\tanh(U_{b,t}^{\mathrm{raw}})\tanh(V_{b,t}^{\mathrm{raw}})^{\top}}{\sqrt{r}} \\
&\quad + \operatorname{diag}\!\left(\tanh(\delta_{b,t}^{\mathrm{raw}})\right), \\
\widetilde A_{b,t} &= \frac{A_{b,t}}{\max\!\left(1,\lVert A_{b,t}\rVert_F\right)}, \quad
g_{b,t} = \sigma\!\left(G_{\eta}(s_{b,t})\right), \\
\eta_{K,b,t} &= \eta_0\left(0.25+0.75g_{b,t}\right), \\
K_{b,t} &= \alpha I+\eta_{K,b,t}\widetilde A_{b,t}.
\end{aligned}
\label{eq:appendix_koopman_operator}
\end{equation}

Here,
\(U_{b,t}^{\mathrm{raw}},V_{b,t}^{\mathrm{raw}}
\in\mathbb{R}^{d_K\times r}\),
\(\delta_{b,t}^{\mathrm{raw}}\in\mathbb{R}^{d_K}\), and
\(r\ll d_K\). We use \(\eta_0=0.05\) and \(\alpha=0.90\), so
\(\eta_{K,b,t}\in(0.0125,0.05)\). Since
\(\lVert\widetilde A_{b,t}\rVert_2\leq1\), the default setting gives
\(\lVert K_{b,t}\rVert_2<0.95\), limiting amplification without
implying dynamical stability. Because the map remains sample-conditioned,
it is not interpreted as a shared physical Koopman operator.

Using the implementation ordering \((V,I,Q)\), define
\begin{equation}
\psi_{b,t}^{-}:=\psi_{b,t}^{V},\qquad
\psi_{b,t}^{0}:=\psi_{b,t}^{I},\qquad
\psi_{b,t}^{+}:=\psi_{b,t}^{Q}.
\label{eq:appendix_channel_roles}
\end{equation}
These symbols specify channel roles rather than temporal states. Two
separate two-layer MLP heads receive the same operator context and
produce a scalar modulation and an element-wise mixing coefficient:
\begin{equation}
\begin{aligned}
d_{b,t}
&=
1+0.10\tanh\!\left(H_d(c_{K,b,t})\right)
\in(0.90,1.10),\\
m_{K,b,t}
&=
0.05\sigma\!\left(H_m(c_{K,b,t})\right)
\in(0,0.05)^{d_K},\\
\epsilon_{b,t}^{K}
&=
\eta_r\tanh\!\left(
R(\operatorname{sg}(\psi_{b,t}^{0}))
\right),
\qquad \eta_r=0.05.
\end{aligned}
\label{eq:appendix_cross_channel_modulation}
\end{equation}
The capacity-channel observable is reconstructed as
\begin{equation}
\widehat{\psi}_{b,t}^{+}
=
d_{b,t}K_{b,t}\psi_{b,t}^{0}
+
m_{K,b,t}\odot\psi_{b,t}^{-}
+
\epsilon_{b,t}^{K}.
\label{eq:appendix_observable_prediction}
\end{equation}
The heads \(H_d\) and \(H_m\) are independent of the operator heads
that generate the low-rank and diagonal components.
The Koopman-inspired cross-channel consistency loss is
\begin{equation}
\mathcal{L}_{\mathrm{cc}}
=
\operatorname{SmoothL1}
\left(
\widehat{\psi}_{b,t}^{+},
\psi_{b,t}^{+}
\right)
+
0.05\left\|
\epsilon_{b,t}^{K}
\right\|_2^2.
\label{eq:appendix_koopman_consistency}
\end{equation}
The ordered channels are not interpreted as successive physical
states, and \(K_{b,t}\) is not treated as a shared physical dynamics
operator. The branch instead provides a Koopman-inspired lifted
consistency regularizer. At inference, prediction follows only the
bounded structural pathway in
Eq.~\eqref{eq:appendix_prediction_path}.

\subsection{Label Normalization and Training Objective}

Each target is standardized independently using only the 1,501
training examples:
\begin{equation}
\widetilde y_{n,j}
=
\frac{
y_{n,j}-\mu_j^{\mathrm{train}}
}{
\max(\sigma_j^{\mathrm{train}},10^{-8})
}.
\label{eq:label_standardization}
\end{equation}
The fitted values are reported in
Table~\ref{tab:label_scaler_stats} and reused unchanged for validation
and test data.

At epoch \(r\), define the standardized error
\[
\varepsilon_{n,j}
=
\widehat{\widetilde y}_{n,j}
-
\widetilde y_{n,j}.
\]
The hybrid loss for output \(j\) is
\begin{equation}
\begin{aligned}
\ell_j
&=
\frac{\lambda_j(r)}
{\sum_n a_{n,j}}
\sum_n
a_{n,j}\mathcal{H}_{n,j},\\
\mathcal{H}_{n,j}
&=
\alpha_r\varepsilon_{n,j}^{2}
+
(1-\alpha_r)
\operatorname{SmoothL1}(\varepsilon_{n,j}),
\end{aligned}
\label{eq:hybrid_loss}
\end{equation}
where \(\alpha_r\) follows a cosine schedule from 0.7 to 0.3.

The seven output losses are macro-averaged. Uniform sample weighting
is used except for CFR. Specifically,
\(\lambda_j=1\) for NDC, SOH, CF, ICR, NCC, and QTY.
For CFR, \(\lambda_{\mathrm{CFR}}=1\) during the first 20\% of
training and then increases linearly to 1.3. CFR samples additionally
receive
\begin{equation}
a_{n,\mathrm{CFR}}
=
1+0.3
\frac{
\operatorname{clip}
(
|\widetilde y_{n,\mathrm{CFR}}|,
0,3
)
}{3}.
\label{eq:cfr_sample_weight}
\end{equation}
The additional standardized-space CFR term is
\begin{equation}
\mathcal{L}_{\mathrm{CFR}}
=
\frac{1}{N}
\sum_{n=1}^{N}
\operatorname{SmoothL1}\!\left(
\widehat{\widetilde y}_{n,\mathrm{CFR}}
-
\widetilde y_{n,\mathrm{CFR}}
\right).
\label{eq:additional_cfr_loss}
\end{equation}

Let \(\widehat y_{n,j}\) denote the prediction mapped back to original
target units. Cross-output consistency is imposed in original target
units by
\begin{equation}
\begin{aligned}
\mathcal{L}_{\mathrm{cross}}
=\frac{1}{N}\sum_{n=1}^{N}
\Bigl[
&\operatorname{SmoothL1}\!\left(
\widehat{\mathrm{SOH}}_n
+
\widehat{\mathrm{CF}}_n
-1
\right)\\
&+\max\!\left(0,-\widehat{\mathrm{CFR}}_n\right)
\Bigr].
\end{aligned}
\label{eq:cross_output_loss}
\end{equation}
The first term penalizes violations of
\(\widehat{\mathrm{SOH}}_n+\widehat{\mathrm{CF}}_n=1\), and the
second penalizes negative CFR predictions. The task objective is
\begin{equation}
\mathcal{L}_{\mathrm{task}}
=
\frac{1}{7}
\sum_{j=1}^{7}\ell_j
+
0.1\mathcal{L}_{\mathrm{CFR}}
+
0.1\mathcal{L}_{\mathrm{cross}}.
\label{eq:implemented_task_objective}
\end{equation}
The complete objective is
\begin{equation}
\mathcal{L}
=
\mathcal{L}_{\mathrm{task}}
+
\lambda_{\mathrm{cc}}\mathcal{L}_{\mathrm{cc}},
\label{eq:implemented_objective}
\end{equation}
with \(\lambda_{\mathrm{cc}}=0.05\).

\begin{table}[t]
\centering

\begin{tabular}{lrr}
\toprule
\textbf{Task} & \(\boldsymbol{\mu_j^{\mathrm{train}}}\) &
\(\boldsymbol{\sigma_j^{\mathrm{train}}}\) \\
\midrule
NDC & 0.9311598 & 0.1476210 \\
SOH & 0.9813603 & 0.0279242 \\
CF  & 0.0186397 & 0.0279242 \\
CFR & \(1.6988796\times10^{-4}\) & \(3.3969091\times10^{-4}\) \\
ICR & 85.042342 & 120.520454 \\
NCC & 0.9296390 & 0.1455620 \\
QTY & 9.4983524 & 17.101401 \\
\bottomrule
\end{tabular}
\caption{Training-label means and standard deviations in original
target units. NDC and NCC correspond to the implementation outputs
named SOC and CA, respectively.}
\label{tab:label_scaler_stats}
\end{table}

\begin{table}[t]
\centering
\setlength{\tabcolsep}{4pt}
\resizebox{\columnwidth}{!}{%
\begin{tabular}{ccccc}
\toprule
\textbf{Masking probability \(p\)}
& \textbf{Runs}
& \textbf{Realized \(p_{\mathrm{pred}}\)}
& \textbf{Overall}
& \(\mathrm{Overall}_{-\mathrm{CFR}}\) \\
\midrule
0.00 & 3  & 0.000 & 0.1242 & 0.0796 \\
0.25 & 30 & 0.264 & 0.1252 & 0.0807 \\
0.50 & 30 & 0.500 & 0.1257 & 0.0813 \\
0.75 & 30 & 0.748 & 0.1265 & 0.0821 \\
1.00 & 3  & 1.000 & 0.1271 & 0.0829 \\
\bottomrule
\end{tabular}}
\caption{Test-time availability stress test for prediction-relevant
structural descriptors. The \(dQ/dV\) lifecycle terrain and
endpoint-derived structural proxy are independently masked at the cell
level with probability \(p\). Realized \(p_{\mathrm{pred}}\) is
computed over these two prediction-relevant descriptors. Material
context is excluded because it is inactive in the inference pathway.
Intermediate probabilities use ten mask seeds for each of three model
seeds; \(p=0\) and \(p=1\) use one deterministic evaluation per model
seed.}
\label{tab:missingness_diagnostic}
\end{table}
\begin{table*}[htbp]
\centering
\resizebox{\textwidth}{!}{%
\begin{tabular}{llrrr}
\toprule
\textbf{Source} & \textbf{Chemistry / form factor} &
\textbf{Candidate cells} & \textbf{Effective cells} &
\textbf{Test windows} \\
\midrule
CALCE & Li-ion, LCO/graphite, prismatic & 13 & 13 & 65 \\
HNEI & Li-ion, LCO+NCM442/graphite, 18650 & 14 & 14 & 70 \\
MATR & Li-ion, LFP/graphite, 18650 & 169 & 169 & 845 \\
\bottomrule
\end{tabular}}
\caption{Held-out sources and evaluable sample counts.}
\label{tab:holdout_source_metadata}
\end{table*}

\begin{table*}[!t]
\centering
\resizebox{\textwidth}{!}{%
\begin{tabular}{llcccccc}
\toprule
& &
\multicolumn{2}{c}{\textbf{TimesNet}} &
\multicolumn{2}{c}{\textbf{DLinear}} &
\multicolumn{2}{c}{\textbf{PatchTST}} \\
\cmidrule(lr){3-4}
\cmidrule(lr){5-6}
\cmidrule(lr){7-8}

\textbf{Task} & \textbf{Metric}
& \textbf{+STEK} & \textbf{Base}
& \textbf{+STEK} & \textbf{Base}
& \textbf{+STEK} & \textbf{Base} \\
\midrule

\multirow{2}{*}{NDC}
& MAE $\downarrow$
& \textbf{0.90 $\pm$ 0.06}
& 0.99 $\pm$ 0.15
& \textbf{1.90 $\pm$ 0.09}
& 2.55 $\pm$ 0.05
& \textbf{1.09 $\pm$ 0.04}
& 1.14 $\pm$ 0.02 \\

& MaxAE $\downarrow$
& \textbf{7.65 $\pm$ 1.15}
& 7.98 $\pm$ 1.38
& \textbf{44.76 $\pm$ 2.02}
& 56.69 $\pm$ 0.83
& 7.83 $\pm$ 0.78
& \textbf{7.27 $\pm$ 1.44} \\

\multirow{2}{*}{SOH}
& MAE $\downarrow$
& \textbf{0.40 $\pm$ 0.03}
& 0.47 $\pm$ 0.02
& \textbf{0.75 $\pm$ 0.03}
& 1.03 $\pm$ 0.03
& \textbf{0.54 $\pm$ 0.06}
& 0.57 $\pm$ 0.02 \\

& RMSE $\downarrow$
& \textbf{0.99 $\pm$ 0.02}
& 1.12 $\pm$ 0.10
& \textbf{1.52 $\pm$ 0.06}
& 1.77 $\pm$ 0.01
& \textbf{1.17 $\pm$ 0.04}
& 1.23 $\pm$ 0.01 \\

\multirow{2}{*}{CF}
& MAE $\downarrow$
& \textbf{0.38 $\pm$ 0.02}
& 0.45 $\pm$ 0.03
& \textbf{0.79 $\pm$ 0.03}
& 1.03 $\pm$ 0.03
& \textbf{0.54 $\pm$ 0.05}
& 0.56 $\pm$ 0.02 \\

& RMSE $\downarrow$
& \textbf{0.97 $\pm$ 0.02}
& 1.10 $\pm$ 0.11
& \textbf{1.55 $\pm$ 0.04}
& 1.78 $\pm$ 0.01
& \textbf{1.17 $\pm$ 0.04}
& 1.22 $\pm$ 0.03 \\

\multirow{2}{*}{CFR}
& MAE $\downarrow$
& \textbf{1.33 $\pm$ 0.00}
& 1.35 $\pm$ 0.01
& \textbf{1.42 $\pm$ 0.02}
& 1.48 $\pm$ 0.01
& \textbf{1.32 $\pm$ 0.02}
& 1.35 $\pm$ 0.03 \\

& RMSE $\downarrow$
& \textbf{3.00 $\pm$ 0.05}
& 3.03 $\pm$ 0.04
& \textbf{3.03 $\pm$ 0.01}
& 3.09 $\pm$ 0.01
& \textbf{2.95 $\pm$ 0.05}
& 3.00 $\pm$ 0.04 \\

\multirow{2}{*}{ICR}
& MAE ($\mathrm{m}\Omega$) $\downarrow$
& 6.54 $\pm$ 0.98
& \textbf{5.46 $\pm$ 0.61}
& \textbf{15.98 $\pm$ 2.27}
& 28.96 $\pm$ 0.65
& \textbf{8.70 $\pm$ 0.35}
& 9.02 $\pm$ 0.75 \\

& RE (\%) $\downarrow$
& 11.99 $\pm$ 2.45
& \textbf{9.09 $\pm$ 1.41}
& \textbf{48.44 $\pm$ 6.08}
& 85.76 $\pm$ 1.92
& \textbf{14.26 $\pm$ 0.81}
& 14.97 $\pm$ 2.52 \\

\multirow{2}{*}{NCC}
& MAE $\downarrow$
& \textbf{0.85 $\pm$ 0.03}
& 0.96 $\pm$ 0.11
& \textbf{1.89 $\pm$ 0.10}
& 2.50 $\pm$ 0.05
& \textbf{1.06 $\pm$ 0.15}
& 1.16 $\pm$ 0.10 \\

& RMSE $\downarrow$
& \textbf{1.31 $\pm$ 0.09}
& 1.43 $\pm$ 0.14
& \textbf{4.86 $\pm$ 0.20}
& 6.10 $\pm$ 0.06
& \textbf{1.61 $\pm$ 0.16}
& 1.66 $\pm$ 0.21 \\

QTY
& RQE (\%) $\downarrow$
& \textbf{6.74 $\pm$ 0.64}
& 8.38 $\pm$ 2.94
& \textbf{58.21 $\pm$ 2.06}
& 86.17 $\pm$ 2.43
& \textbf{13.72 $\pm$ 2.71}
& 18.24 $\pm$ 8.00 \\

\midrule
OVERALL
& Score $\downarrow$
& \textbf{12.42 $\pm$ 0.30}
& 13.29 $\pm$ 0.71
& \textbf{21.60 $\pm$ 0.86}
& 28.48 $\pm$ 0.14
& \textbf{14.81 $\pm$ 0.66}
& 15.51 $\pm$ 0.54 \\
\bottomrule
\end{tabular}}
\caption{Generality of STEK across temporal backbones on
MIX-battery. NDC, SOH, CF, NCC, and Overall are multiplied
by $10^2$, while CFR is multiplied by $10^4$.
$\downarrow$ indicates that lower values are better.
For each backbone, the better result between the base model
and its STEK-enhanced variant is bolded.}
\label{tab:mix_battery_comparison_ablation}
\end{table*}

\begin{table}[t]
\centering
\small
\setlength{\tabcolsep}{3.2pt}
\begin{tabular}{lcccc}
\toprule
\textbf{Model}
& \textbf{Params}
& \textbf{FLOPs}
& \textbf{Memory}
& \textbf{Overall $\downarrow$} \\
\midrule
MambaLithium  & 0.09  & 0.02   & 14.93 & $16.24 \pm 0.59$ \\
LiPM          & 0.66  & 0.12   & 13.64 & $22.11 \pm 1.10$ \\
CPTransformer & 0.12  & 0.01   & 17.63 & $16.51 \pm 1.29$ \\
iTransformer  & 0.78  & 0.001  & 12.70 & $15.08 \pm 0.47$ \\
Koopa         & 0.09  & 0.001  & 12.81 & $16.27 \pm 0.45$ \\
KNF           & 0.23  & 0.01   & 10.64 & $17.26 \pm 0.38$ \\
DLinear       & 0.02  & 0.0001 & 9.68  & $28.48 \pm 0.14$ \\
STEK-DLinear  & 2.23  & 0.31   & 20.71 & $21.60 \pm 0.86$ \\
PatchTST      & 5.28  & 0.01   & 39.72 & $15.51 \pm 0.54$ \\
STEK-PatchTST & 7.53  & 0.32   & 51.20 & $14.81 \pm 0.66$ \\
TimesNet      & 10.06 & 93.72  & 72.47 & $13.29 \pm 0.71$ \\
STEK-TimesNet & 11.58 & 94.02  & 77.75 & $\mathbf{12.42 \pm 0.30}$ \\
\bottomrule
\end{tabular}
\caption{Model efficiency under the MIX-battery seven-output setting.
Parameters, FLOPs, and memory are reported in M, G, and MB,
respectively. Parameter counts include all trainable modules, including
the training-only Koopman-inspired branch. FLOPs measure one FP16
prediction forward pass and exclude offline lifecycle-terrain
construction, precomputed material-text embeddings, training losses,
backward propagation, and optimizer updates. Values are reported as
mean \(\pm\) standard deviation over three seeds.}
\label{tab:efficiency}
\end{table}

\subsection{Evaluation Metrics}

For predictions \(\{\widehat y_i\}_{i=1}^{N}\) and targets
\(\{y_i\}_{i=1}^{N}\), the standard error metrics are
\begin{align}
\mathrm{MAE}
&=\frac{1}{N}\sum_{i=1}^{N}|y_i-\widehat y_i|,\\
\mathrm{MaxAE}
&=\max_i|y_i-\widehat y_i|,\\
\mathrm{RMSE}
&=\left[
\frac{1}{N}\sum_{i=1}^{N}
(y_i-\widehat y_i)^2
\right]^{1/2}.
\end{align}
NDC uses MAE and MaxAE; SOH, CF, CFR, and NCC use MAE and
RMSE. ICR is additionally evaluated using
\begin{equation}
\mathrm{RE}(\%)
=\frac{100}{N}\sum_{i=1}^{N}
\frac{|y_i-\widehat y_i|}{|y_i|+\epsilon},
\end{equation}
and QTY uses
\begin{equation}
\mathrm{RQE}(\%)
=100\,
\frac{\sum_{i=1}^{N}|y_i-\widehat y_i|}
{\sum_{i=1}^{N}|y_i|+\epsilon}.
\end{equation}
Lower values are better for all metrics.

For aggregate reporting, predictions and labels are first returned to
their original units, and
\begin{align}
\mathrm{NMAE}_j
&=\frac{\mathrm{MAE}_j}
{\max(\sigma_j^{\mathrm{train}},10^{-8})},
\label{eq:nmae_appendix}\\
\mathrm{Overall}
&=\frac{1}{7}\sum_{j=1}^{7}\mathrm{NMAE}_j.
\label{eq:overall_appendix}
\end{align}
Thus each predefined output has reporting weight \(1/7\), independent
of its unit or sample count. NDC and NCC are directly recoverable, and
CF is algebraically determined by SOH. Overall is consequently a
benchmark-specific seven-output composite rather than an average over
seven independent latent states. It is reported for compatibility with
the fixed protocol rather than introduced as a post-hoc independent
target metric. It is interpreted jointly with the per-output results,
the deterministic capacity-recovery diagnostic, and the information
groups in
Table~\ref{tab:target_information_groups}.

\(\mathrm{Overall}_{-\mathrm{CFR}}\) is a six-output sensitivity
diagnostic used only in the availability analysis; it is not used for
the main model ranking.

\section{Evaluation Protocol and Additional Results}
\label{app:evaluation_protocol}

\subsection{Matched Baseline Protocol}

Endpoint-only neural baselines share the cell split, endpoint range,
curve input, seven-output objective, training budget, and
validation-based checkpoint selection. MambaLithium and LiPM are
adapted to the canonical endpoint-curve input. These baselines do not
receive archived lifecycle or material-text context.

STEK and the direct-concatenation control receive the same auxiliary
descriptors: material-text metadata, the \(dQ/dV\) lifecycle terrain,
and the endpoint-derived structural proxy. The concatenation control
therefore provides a matched-input comparison of fusion strategies,
although exact parameter matching is not imposed.

\subsection{Availability of Prediction-Relevant Structural Descriptors}

The \(dQ/dV\) terrain and endpoint-derived structural proxy are masked
independently at the cell level without retraining. All endpoint
samples from the same cell share the same descriptor mask. The
realized missing rate
\(p_{\mathrm{pred}}\) is computed only over these two descriptors
because they are the auxiliary inputs that enter the prediction path.
Material context is excluded from \(p_{\mathrm{pred}}\) because it
conditions only the training-time regularizer and is inactive in the
inference pathway. Intermediate probabilities use ten mask seeds
shared across the three model seeds, whereas \(p=0\) and \(p=1\)
require one deterministic evaluation per model seed.

As reported in Table~\ref{tab:missingness_diagnostic}, Overall
increases from \(0.1242\) to \(0.1271\) when both prediction-relevant
structural descriptors are unavailable, corresponding to a \(2.3\%\)
relative increase. The CFR-excluded diagnostic increases from
\(0.0796\) to \(0.0829\), or \(4.1\%\).

The monotonic aggregate degradation indicates that the structural
descriptors provide useful auxiliary information. It does not, by
itself, establish that the two descriptors are mutually complementary.
The bounded modulation and validity-gated residual path retain an
endpoint-curve fallback when the descriptors are absent. This
experiment evaluates descriptor absence rather than noisy, corrupted,
or misleading inputs.

\subsection{Leave-One-Source-Out Protocol}

The matched TimesNet comparisons separately hold out CALCE,
HNEI, and MATR,
which cover distinct cathode families, form factors, source sizes,
and cycling protocols. Table~\ref{tab:holdout_source_metadata}
reports the corresponding evaluable cells and test windows.

All holdout experiments follow the same preprocessing and
cell-level separation protocol. Their task-level results are reported
in the main paper.

\subsection{Detailed Results across Temporal Backbones}

Table~\ref{tab:mix_battery_comparison_ablation} complements Fig.~3
of the main paper with task-level results for TimesNet, DLinear, and
PatchTST. Each backbone and its STEK-enhanced variant use the same
split, training objective, and evaluation protocol.

STEK lowers the fixed-protocol Overall score for all three backbones.
The largest reduction is observed for DLinear, while the TimesNet and
PatchTST results show that the framework is not tied to one temporal
architecture. All seven primary metrics improve for DLinear and
PatchTST; for TimesNet, six primary metrics improve while ICR MAE
degrades. Secondary metrics are not uniformly improved: STEK also
increases the PatchTST NDC MaxAE despite reducing the aggregate score.

\subsection{Efficiency Analysis}

Table~\ref{tab:efficiency} compares model complexity and aggregate
performance under the same FP16 inference setting. Parameter counts
include the temporal backbone, auxiliary encoders, structural
modulation and prediction modules, task heads, and training-only
Koopman-inspired branch.

Lifecycle terrains and MolT5 material-text embeddings are constructed
offline and are excluded from the reported inference FLOPs. The frozen
MolT5 encoder is also excluded from parameter counts. STEK reduces the
fixed-protocol Overall score across all three evaluated backbones while
increasing model size, inference computation, and memory. For
TimesNet, Overall decreases from \(13.29\pm0.71\) to
\(12.42\pm0.30\), illustrating the resulting
accuracy--efficiency trade-off.

\bibliography{references}